\documentclass[journal]{IEEEtran}

\usepackage{times} 
\usepackage{epsfig}
\usepackage{graphicx}
\usepackage{amsmath}
\usepackage{amssymb}
\usepackage{mathrsfs}
\usepackage{overpic}
\usepackage{color}
\usepackage{balance}
\usepackage{booktabs}
\usepackage{enumerate}
\usepackage[linesnumbered,ruled]{algorithm2e}
\usepackage[square,sort,comma,numbers]{natbib}

\begin{document}

\title{Single Image Dehazing Using Ranking Convolutional Neural Network}

\author{Yafei~Song,~
        Jia~Li,~
        Xiaogang~Wang~
        and~Xiaowu~Chen
\thanks{This paper was published on IEEE Transaction on Multimedia, vol. 20, No. 6, pp: 1548-1560. \emph{(Corresponding author: Xiaowu Chen (e-mail: chen@buaa.edu.cn).)} }
\thanks{Y. Song, X. Wang and X. Chen are with the State Key Laboratory of Virtual Reality Technology and Systems, School of Computer Science and Engineering, Beihang University, Beijing 100191, China. (e-mail: songyf@buaa.edu.cn; wangxiaogang@buaa.edu.cn; chen@buaa.edu.cn)}

\thanks{J. Li is with the State Key Laboratory of Virtual Reality Technology and Systems, School of Computer Science and Engineering, Beihang University, Beijing 100191, China, and also with the International Research Institute for Multidisciplinary Science, Beihang University, Beijing 100191, China. (e-mail: jiali@buaa.edu.cn)}
}


\maketitle

\begin{abstract}
Single image dehazing, which aims to recover the clear image solely from an input hazy or foggy image, is a challenging ill-posed problem. Analysing existing approaches, the common key step is to estimate the haze density of each pixel. To this end, various approaches often \emph{heuristically designed} haze-relevant features. Several recent works also automatically learn the features via directly exploiting Convolutional Neural Networks (CNN). However, it may be insufficient to fully capture the intrinsic attributes of hazy images. To obtain effective features for single image dehazing, this paper presents a novel Ranking Convolutional Neural Network (Ranking-CNN). In Ranking-CNN, a novel ranking layer is proposed to extend the structure of CNN so that the statistical and structural attributes of hazy images can be simultaneously captured. By training Ranking-CNN in a well-designed manner, powerful haze-relevant features can be \emph{automatically learned} from massive hazy image patches. Based on these features, haze can be effectively removed by using a haze density prediction model trained through the random forest regression. Experimental results show that our approach outperforms several previous dehazing approaches on synthetic and real-world benchmark images. Comprehensive analyses are also conducted to interpret the proposed Ranking-CNN from both the theoretical and experimental aspects.
\end{abstract}


\IEEEpeerreviewmaketitle

\section{Introduction}
\IEEEPARstart{I}{n} real-world scenarios, small particles suspending in the atmosphere ({\em e.g.}, droplets and dusts) often scatter the light. As a consequence, the clarity of an image would be seriously degraded, which may decrease the performance of many multi-media processing systems, {\em e.g.}, content-based image retrieval \cite{Gao:2016:TMM}. Image enhancement methods \cite{Wang:2016:TMM, Xu:2014:TMM} can only alleviate this problem slightly. It is still helpful to develop effective dehazing methods to recover the clear image from an input hazy or foggy image.

\begin{figure}[t]
  \centering
  \includegraphics[width = 84mm]{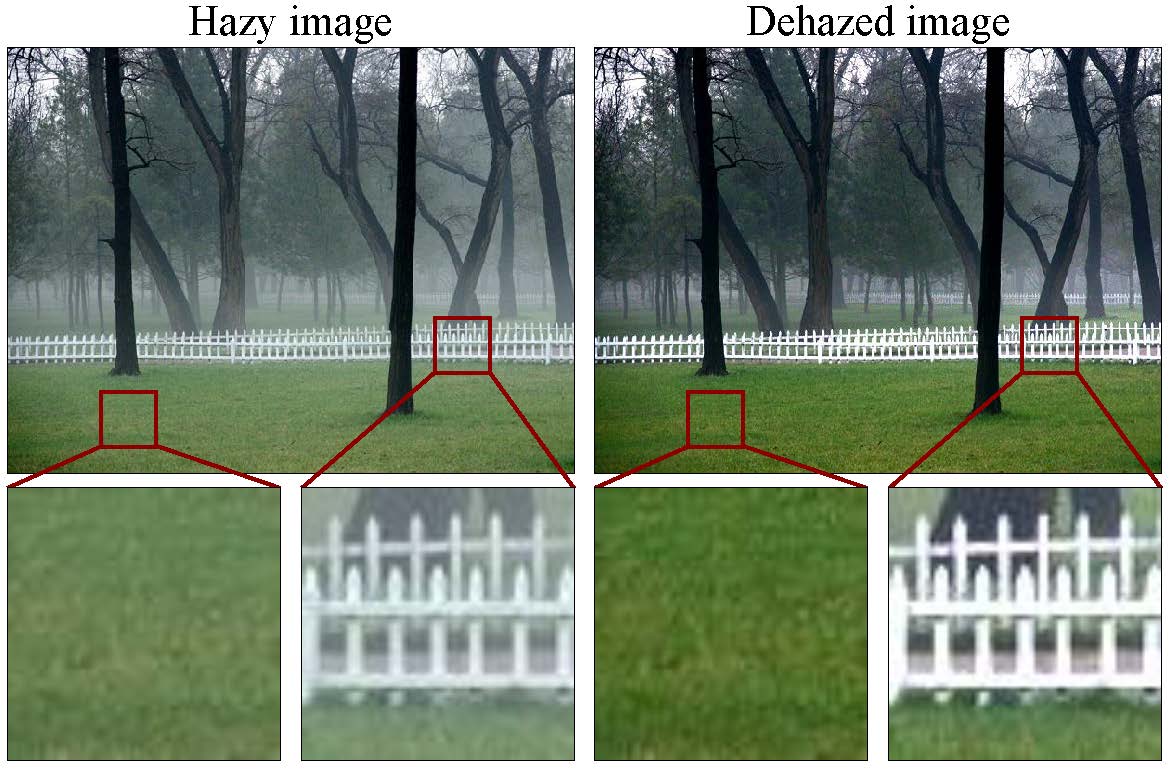}
\caption{Both statistical and structural attributes of image patches are useful for dehazing. For example, the grass patches can be dehazed according to their color statistics ({\em i.e.}, the statistical attributes of image patches), while the haze over fence can be removed according to their gradients ({\em i.e.}, the structural attributes of image patches).}
\label{fig:motivation}
\end{figure}

In the past decades, the problem of haze formation has been extensively studied in atmospheric optics \cite{Timofeev:2008:BOOK}. It is widely acknowledged that a hazy image can be regarded as a convex combination of scene radiance and atmospheric light \cite{He:2009:CVPR, Tang:2014:CVPR, Fattal:2014:TOG, Berman:2016:CVPR, Ren:2016:ECCV, Cai:2016:TIP, Wang:2017:TMM}. The combination coefficient is often called the \textit{transmission}. As a result, the task of image dehazing can be formulated as \textit{recovering the scene radiance from a hazy image by estimating the atmospheric light and the transmission}. 

Under this formulation, two kinds of dehazing approaches have been proposed in the literature. Some of them propose to dehaze an image under the assistance of additional information, {\em e.g.}, scene depth \cite{Kopf:2008:TOG}, images taken under different weathers \cite{Nayar:1999:ICCV,Narasimhan:2000:CVPR}. However, such additional information may not be always available, which prevents the further usage of these dehazing approaches in many real-world scenarios. On the contrary, some approaches propose to directly dehaze a single image, which is an ill-posed problem since the atmospheric light and the transmission need to be simultaneously recovered for each image pixel. To address this issue, these approaches often assume that the atmospheric light is constant for every pixel in one input image, so that it can be estimated first in a pre-processing step. After that, the dehazing process can be simplified as a transmission estimation problem. For instance, He {\em et al.} \cite{He:2009:CVPR} propose the dark channel prior which is proved to be effective in transmission estimation. Tang {\em et al.} \cite{Tang:2014:CVPR} incorporate four types of features to train a regression model for transmission prediction. Fattal \cite{Fattal:2014:TOG} utilizes local color-lines prior in clear images to estimate the transmission. Berman {\em et al.} \cite{Berman:2016:CVPR} further propose non-local haze-line prior. In many cases, these approaches achieve impressive performance. However, for each prior, there are often images which may not meet it. Therefore, the heuristic designed priors (or features) may be insufficient to fully capture the intrinsic attributes of hazy images.

Inspired by the impressive success of Convolutional Neural Networks (CNN) \cite{LeCun:1998:IEEE}, \textit{e.g.}, image classification/annotation \cite{Krizhevsky:2012:NIPS, Wu:2015:TBD}, object detection \cite{Erhan:2014:CVPR}, semantic segmentation \cite{Farabet:2013:TPAMI}, and image denoising \cite{Xie:2012:NIPS,Agostinelli:2013:NIPS}, this paper prefers to automatically learn the haze-relevant features from massive hazy images. Two recent works \cite{Ren:2016:ECCV, Cai:2016:TIP} also hold the same basic idea and adopt CNN to perform image dehazing. Ren {\em et al.} \cite{Ren:2016:ECCV} directly estimate the whole transmission map from an input image under the multi-scale FCN (fully convolutional networks) framework \cite{Long_2015_CVPR}. Cai {\em et al.} \cite{Cai:2016:TIP} use a regression network to estimate the transmission of each pixel from its local surrounding patch. However, these two works mainly exploit existing layers to construct their CNNs. In contrast, we propose a new layer, named ranking layer, derived from our insight on this problem, which can facilitate the learning process of haze-relevant features.

By analysing the mechanism of existing image dehazing methods, we find that statistical attributes are essential, \textit{e.g.}, dark channel prior \cite{He:2009:CVPR}, haze-line prior \cite{Berman:2016:CVPR} and color-lines prior \cite{Fattal:2014:TOG}. But the classical CNN, while capturing the structural attributes well ({\em e.g.}, the fence in Fig.~\ref{fig:motivation}), may lack the ability to capture the statistical attributes ({\em e.g.}, the grass in Fig.~\ref{fig:motivation}). To alleviate this problem, we propose a novel ranking layer which can be embedded in the structure of classical CNN to form the Ranking-CNN. A Ranking-CNN can capture the structural and statistical attributes simultaneously. As a straightforward method, an end-to-end regression network can be established to estimate the transmission of each pixel from its surrounding local patch. However, since the regression target is only a real value between $(0, 1]$, when training the network using backward propagation algorithm, the gradient may be small and not robust. Therefore, it is difficult to effectively train the deep network. To this end, the regression problem is converted into a classification problem. Then the Ranking-CNN can be effectively trained on massive hazy image patches, and various types of haze-relevant features can be automatically learned. Based on these features, the random forest is further adopted to train a regression model so as to predict the transmission. Experimental results on plentiful synthetic and real-world images show that the proposed approach outperforms several previous outstanding approaches. 

The main contributions of this paper include: First, we propose a novel ranking layer as well as its forward and backward computations, and theoretical analyses illuminate its excellent ability to capture statistical attributes. Second, by incorporating the ranking layer into the classical CNN, we construct a Ranking-CNN to learn effective haze-relevant features, which demonstrates impressive performance in image dehazing. Third, we benchmark the proposed dehazing approach and several state-of-the-art methods on extensive qualitative and quantitative experiments, in which the proposed approach achieves satisfactory performance.

The rest of this paper is organized as follows. Section~\ref{sect:related:work} presents some related works. Section~\ref{sect:problem:formulation} formulates the problem and overviews our pipeline. Then each step is detailedly explained in Section~\ref{sect:Ranking-CNN}. Finally we show the experimental results in Section~\ref{sect:experiments} and conclude this paper in Section~\ref{sect:conclusion}.

\section{Related Work}
\label{sect:related:work}
In the past two centuries, the interaction phonomenon of light with the atmosphere has been widely studied \cite{Middleton:1957:BOOK, Mccartney:1976:Optics, Timofeev:2008:BOOK}, which is known as atmospheric optics. Based on the physical phonomenon, depending on whether using additional information, there are mainly two kinds of image dehazing methods. We then review the related works from this perspective. In addition, we also briefly introduce several representative works on deep neural network.

\textbf{Image dehazing with additional information}. Early methods usually use additional information to dehaze images. Nayar and Narasihan \cite{Nayar:1999:ICCV, Narasimhan:2000:CVPR} restore the scene structure from multiple images captured under different weather conditions, then the clear image can be recovered. Schechner {\em et al.} \cite{Schechner:2001:CVPR} observe that the scattered atmospheric light is usually partially polarized, then they take two or more images through a polarizer at different orientations for image dehazing.  Shwartz {\em et al.} \cite{Shwartz:2006:CVPR} automatically recover the parameters of the atmospheric light needed by polarizer based image dehazing methods. Kopf {\em et al.} \cite{Kopf:2008:TOG} use the geometry of the scene to dehaze image via registering the hazy image into 3D scenes manually. However, as these additional information is usually difficult to obtain, these methods have many limitations.

\begin{figure*}[!t]
  \centering
  \includegraphics[width = 160mm]{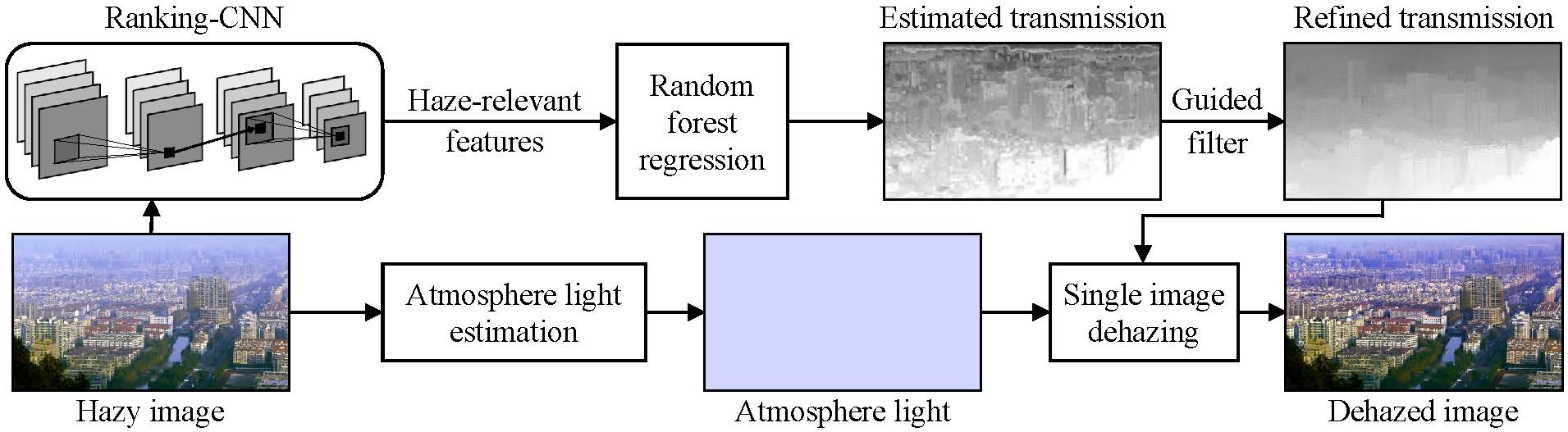}
\caption{System framework of our approach. Given a hazy image, we first estimate a global atmospheric light and use a pre-trained Ranking-CNN to extract haze-relevant features for each pixel from its surrounding patch. After that, the initial transmission is estimated via a random forest regression model, which is then refined through a guided filter. Finally, the clear image is recovered through single image dehazing.}
\label{fig:overview}
\end{figure*}

\textbf{Single image dehazing}. As single image dehazing is an ill-posed problem, various priors and hypotheses have been proposed to tackle this problem. Oakley and Bu {\em et al.} \cite{Oakley:2007:TIP} assume a constant air-light and estimate it via finding the minimum of a global cost function. Tan \cite{Tan:2008:CVPR} removes the haze layer based on the observations that clear images have more contrast and the transmission tends to be smooth. Fattal \cite{Fattal:2008:TOG} assumes that the shading and transmission functions are locally statistically uncorrelated. Tarel and Hauti\`ere \cite{Tarel:2009:ICCV} propose a fast algorithm whose complexity is a linear function of the image size. Kratz and Nishino \cite{Kratz:2009:ICCV} assume that the albedo and depth are statistically independent, then formulate a factorial Markov random field to estimate the transmission. He {\em et al.} \cite{He:2009:CVPR} observe that the lowest value of each channel in a local image patch tends to be zero for clear images, which called dark channel prior. Wen {\em et al.} \cite{Wen:2013:ISCAS} further develop the underwater dark channel prior for image enhancement. Gibson {\em et al.} \cite{Gibson:2012:TIP} investigate the dehazing effects on image and video coding. They further \cite{Gibson:2013:ICIP} use locally adaptive Wiener filter to refine the estimated density of haze. Yan {\em et al.} \cite{Yan:2013:SIGATB} reduce the amplified noise in the dehazed result image restored from dense haze. Fattal \cite{Fattal:2014:TOG} utilizes the color-lines prior in local image patch. Sulami {\em et al.} \cite{Sulami:2014:ICCP} apply the color-lines prior to estimate an appropriate global constant atmospheric light vector. Wang and Fan \cite{Wang:2014:TIP} propose a multiscale depth fusion (MDF) method with local Markov regularization to blend multi-level details of chromaticity priors. Zhu {\em et al.} \cite{Zhu:2015:TIP} propose a color attenuation prior and further apply a linear model for haze removal. Wang {\em et al.} \cite{Wang:2017:TMM} propose a fast method based on linear transformation. For each prior, it can be applied to a range of hazy images, however, there are often images which may not meet it. To this end, this paper aims at automatically learning information from massive data.

Recently, there are several learning-based image dehazing methods. Tang {\em et al.} \cite{Tang:2014:CVPR} train a regression model to estimate the transmission via incorporating four types of haze-relevant features. Two recent works \cite{Ren:2016:ECCV, Cai:2016:TIP} also adopt CNN to perform image dehazing. Ren {\em et al.} \cite{Ren:2016:ECCV} directly estimate the whole transmission map from an input image via multi-scale CNN under the FCN framework \cite{Long_2015_CVPR}. Cai {\em et al.} \cite{Cai:2016:TIP} use a regression network to estimate the transmission of each pixel from its surrounding patch. However, these works mainly exploit existing hand-crafted features or classical CNNs. In contrast, we propose a novel Ranking-CNN to simultaneously capture statistical and structure attributes, which both are essential for single image dehazing.

\textbf{Deep neural networks}. Deep neural networks, also well known as deep learning or feature learning, are more powerful than shallow learning algorithms \cite{Hinton:2006:NC}. Many researchers use deep learning to perform high level computer vision tasks and significantly improve the performance, such as image classification \cite{Krizhevsky:2012:NIPS, He:2016:CVPR}, object detection \cite{Erhan:2014:CVPR, Szegedy:2015:CVPR}, and semantic labelling \cite{Farabet:2013:TPAMI, Long_2015_CVPR, Chen:2015:ICLR}. Researches also have applied deep neural network to tackle low level problems and obtain promising results. Xie {\em et al.} \cite{Xie:2012:NIPS} propose the Stacked Sparse Denoising Auto-encoders (SSDA) to perform image denoising and inpainting. Agostinelli {\em et al.} \cite{Agostinelli:2013:NIPS} further propose adaptive multi-column stacked sparse denoising autoencoder (AMC-SSDA) to tackle multiple types of noise. Schuler {\em et al.} \cite{Schuler:2013:CVPR} train a multi-layer perceptron to perform image deconvolution task and obtain satisfactory results. Cho {\em et al.} \cite{Cho:2016:ECCV} applies CNN on image matting. And Shen {\em et al.} \cite{Shen:2016:ECCV} focus on portrait matting. These works demonstrate that the deep neural network can achieve satisfactory results not only on high-level problems but also low-level problems.

\section{Overview}
\label{sect:problem:formulation}

To dehaze an image, we first briefly formulate the formation process of a hazy image. Under the hazy or foggy weather, the scene radiance is scattered by the small particles suspending in the atmosphere. With increasing scene depth, the camera sensor captures less scene radiance but more atmosphere light. Thus, the formation of a hazy image can be described as a convex combination of the scene radiance $\textbf{J}$ and the atmospheric light $\textbf{A}$, which can be formulated as \cite{Nayar:1999:ICCV}
\begin{equation}
  \textbf{I}\left(x\right) = \textbf{J}\left(x\right) t\left(x\right) + \textbf{A}\left(x\right) \left(1-t\left(x\right)\right),
\label{eq:haze:formation}
\end{equation}
where $\textbf{I}\left(x\right)$ is a pixel from the hazy image $\textbf{I}$ and $t\left(x\right)$ is its transmission. As a consequence, the problem of single image dehazing can be described as recovering the scene radiance $\textbf{J}\left(x\right)$ from the hazy pixel $\textbf{I}\left(x\right)$. From \eqref{eq:haze:formation}, we have
\begin{equation}
  \textbf{J}\left(x\right) = \frac{\textbf{I}\left(x\right) - \textbf{A}\left(x\right)\left(1-t\left(x\right)\right)}{t\left(x\right)}.
\label{eq:dehaze}
\end{equation}
Note that the dehazing process in \eqref{eq:dehaze} is ideal and may require slight variations in building the computational model for image dehazing. From \eqref{eq:dehaze}, we find that the dehazing problem can be decomposed to three subproblems, including:
 \begin{enumerate}
 \setlength{\itemsep}{0.5ex}
 \item[1)] Estimate the atmospheric light $\textbf{A}\left(x\right)$,
 \item[2)] Predict the transmission $t\left(x\right)$,
 \item[3)] Recover scene radiance $\textbf{J}\left(x\right)$.
 \end{enumerate}

To address these subproblems, the system framework of our approach is shown in Fig.~\ref{fig:overview}.  Specifically, since transmission prediction is often considered to be the key and most challenging subproblem in image dehazing \cite{He:2009:CVPR, Tang:2014:CVPR, Fattal:2014:TOG}, we propose the Ranking-CNN for this subproblem. Similar to the solutions in \cite{He:2009:CVPR, Tang:2014:CVPR}, we also assume that the atmospheric light is constant for all image pixels. Then, we calculate the dark channel of the input hazy image using the approach in \cite{He:2009:CVPR}, and the atmospheric light $\textbf{A}(x)$ at any pixel $x$ is estimated by averaging the RGB color of the $0.1\%$ pixels with the largest dark channel values.

Once the atmospheric light is estimated, we only have to focus on predicting the transmission $t\left(x\right)$ for every pixel according to its local features. To extract haze-relevant features, the proposed Ranking-CNN extends the structure of the classical CNN by adding a novel ranking layer so that the statistical and structural attributes of hazy image patches can be simultaneously captured. Based on the haze-relevant features, a transmission prediction model is then trained using the random forest regressor. The random forest regressor is adopted due to its several advantages, such as it can measure the importance of features and avoid seriously over-fitting. This regression model can be used to obtain the initial transmission for every pixel in the input image. To avoid edge artifacts, a guided filter is applied to refine the initial transmission, and the refined transmission is combined with the estimated global atmospheric light for image dehazing. As the Ranking-CNN model and the regression model are trained on massive amounts of data, they are effective for different input hazy images. Thus, we only need to train one unique Ranking-CNN model and one unique regression model, which are then used to dehaze any input hazy image.

\section{The Approach}
\label{sect:Ranking-CNN}
In this Section, we first introduce what the ranking layer is and how to add it to the structure of the classical CNN so as to construct the Ranking-CNN. After that, we describe the implementation details of the Ranking-CNN and show how to learn haze-relevant features. Finally, we demonstrate how to dehaze an input image with the features extracted by the Ranking-CNN.

\subsection{Ranking Layer}
\label{sect:ranking:layer}

By analysing the mechanism of existing image dehazing methods, we find that two types of attributes may influence the performance of transmission estimation, including statistical attributes ({\em e.g.}, dark channel prior \cite{He:2009:CVPR} and color-lines prior \cite{Fattal:2014:TOG}) and structural attributes ({\em e.g.}, boundaries \cite{Meng:2013:ICCV}). Inspired by this observation, we propose to automatically learn haze-relevant features through CNN so as to simultaneously capture these two types of attributes. However, CNN performs impressively on capturing the structural attributes due to the usage of convolutional layers, while it often lacks the ability to extract statistical attributes. Thus it is necessary to modify the structure of the classical CNN so as to enhance its ability in extracting haze-relevant features. Toward this end, we propose to add a ranking layer to the classical CNN so as to construct a novel Ranking-CNN.

\begin{figure}[t]
  \centering
  \includegraphics[width = 84mm]{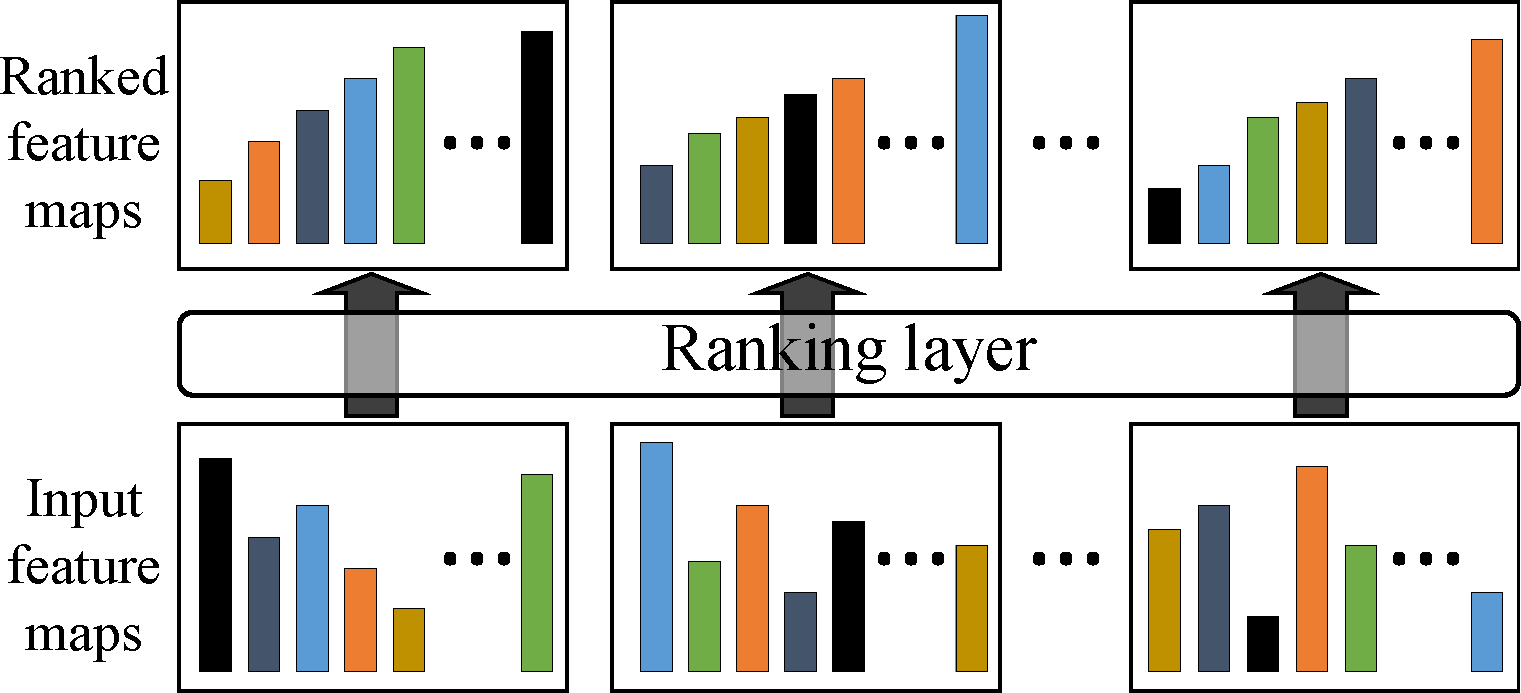}
  \caption{A ranking layer operates separately on each input feature map and only changes the ordering of elements in each feature map other than modifying their values. Note that a feature map is actually a 2D matrix and here we turn them into a 1D vector by sampling elements column-wise so as to provide a better viewing experience.}
\label{fig:ranking:layer}
\end{figure}

\begin{figure}[t]
  \centering
  \includegraphics[width = 84mm]{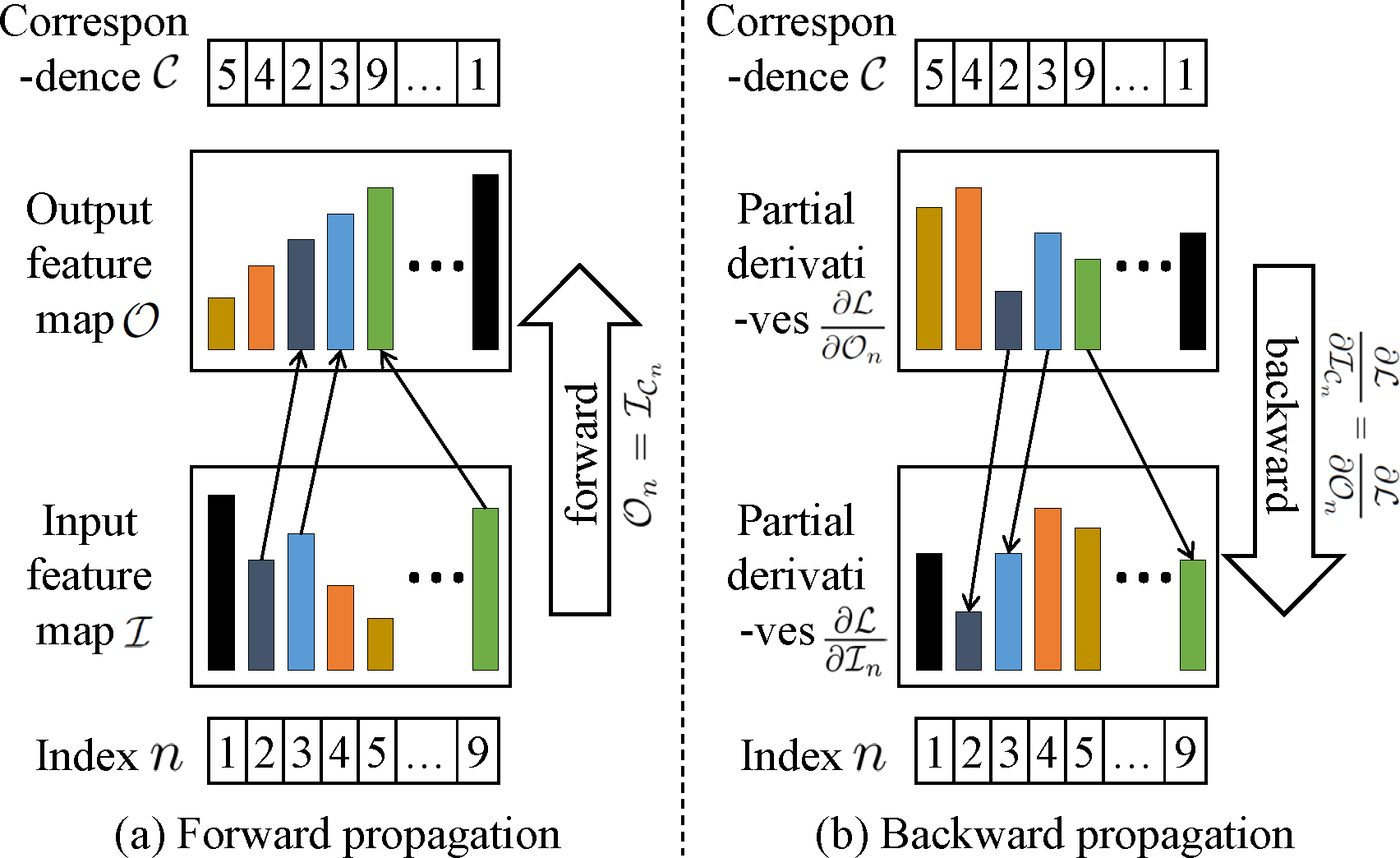}
  \caption{The forward and backward propagation of the ranking layer on a specific feature map. In the forward propagation: the ranking layer sorts all the elements in a feature map and records the correspondence $\mathcal{C}$ between the input and output feature maps. In the backward propagation: the ranking layer propagates the partial derivatives from the output feature map to the input feature map according to the correspondence $\mathcal{C}$.}
\label{fig:ranking:layer:forward:backward}
\end{figure}

For a ranking layer, its input consist of a number of feature maps, which is the same as a common layer of classical CNNs. The proposed ranking layer retains the values of all the elements in a feature map and only changes their ordering. The input of a ranking layer consists of a set of feature maps, and the ranking layer operates separately on each input feature map and output a ranked feature map with the same dimension (as shown in Fig.~\ref{fig:ranking:layer}). Let $\mathcal{I}$ be an input feature map with $N$ elements and $\mathcal{O}$ be its ranked version, we denote the $n$th element of $\mathcal{I}$ and $\mathcal{O}$ as $\mathcal{I}_n$ and $\mathcal{O}_n$, respectively. As shown in Fig.~\ref{fig:ranking:layer:forward:backward} (a), in the forward propagation of a ranking layer, the element $\mathcal{O}_n$ corresponds to the $n$th smallest element in $\mathcal{I}$, whose index is denoted as $\mathcal{C}_n$, \textit{i.e.}, $\mathcal{O}_n = \mathcal{I}_{\mathcal{C}_n}$. To facilitate the operations in the backward propagation, we record such pair-wise correspondences between the elements of input and output feature maps as $\{(\mathcal{C}_n, n)|1\leq{}n\leq{}N\}$.

\begin{figure*}[!t]
  \centering
  \includegraphics[width = 158mm]{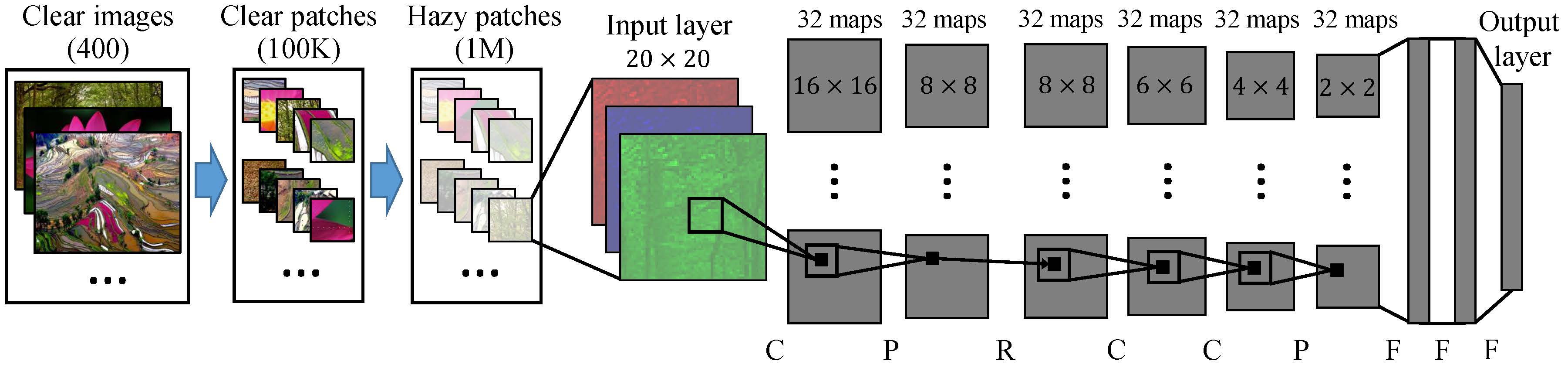}
  \caption{The generation of training data and the structure of the Ranking-CNN. One million training patches are synthesized via adding random haze to 100k clear image patches sampled from 400 clear images. The Ranking-CNN is constructed by adding a ranking layer to the structure of classical CNN (C: convolution; P: max pooling; R: ranking; F: fully-connected).}
\label{fig:Ranking-CNN:structure}
\end{figure*}

\begin{figure}[t]
  \centering
  \includegraphics[width = 84mm]{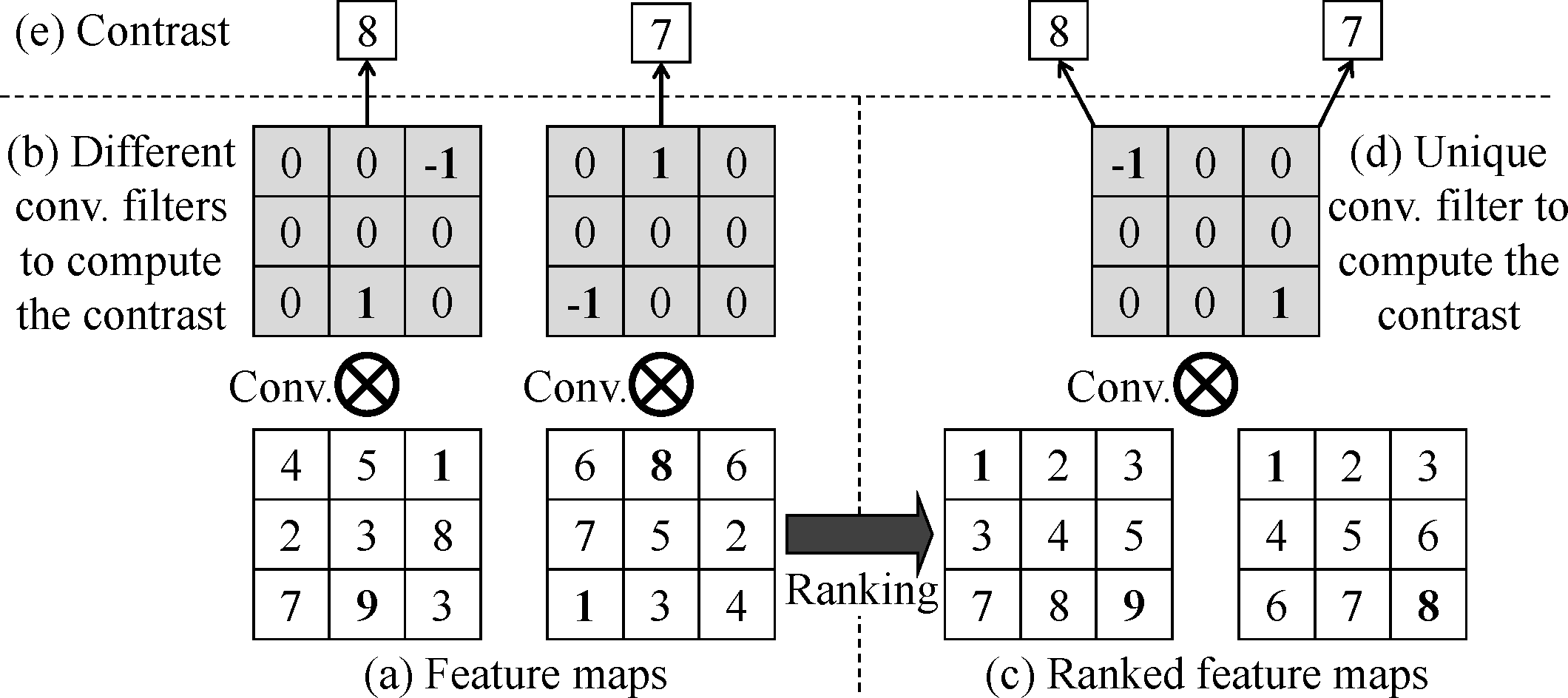}
  \caption{An example to show how the ranking layer facilitates the statistical attributes extraction, \textit{e.g.}, the contrast. For (a) various feature maps, classical CNN may need (b) different convolutional filters to compute the contrast. However, if (c) the feature maps are ranked, only (d) one unique filter is needed.}
\label{fig:ranking:layer:example}
\end{figure}

Based on the pair-wise correspondences $\{(\mathcal{C}_n, n)|1\leq{}n\leq{}N\}$ between the elements of an input feature map and its output ranked version, the backward propagation at the ranking layer can be conducted. As the ranking layer only changes the ordering of elements in each feature map, the partial derivatives of the loss function $\mathcal{L}$ with respect to each output feature $\mathcal{O}_n$ can be directly passed to its corresponding input feature $\mathcal{I}_{C_n}$ as
\begin{equation}
\setcounter{equation}{3}
\frac{\partial{}\mathcal{L}}{\partial \mathcal{I}_{\mathcal{C}_n}} = \frac{\partial \mathcal{L}}{\partial \mathcal{O}_n}.
\label{eq:ranking:backward}
\end{equation}
In Fig.~\ref{fig:ranking:layer:forward:backward} (b), we pick a specific feature map and visually explain the backward propagation of the ranking layer. Note that the ranking layer is parameter-free. No parameter needs to be learned in the backward propagation other than passing the derivatives.

The ranking layer operates separately on each feature map and sorts the elements in an input feature map in ascending order. Since the output feature map is ordered, extracting its statistical attributes, \textit{e.g.}, its contrast, becomes easier. As shown in Fig. \ref{fig:ranking:layer:example}(a), for various feature maps, classical CNN may need different convolutional filters (Fig.~\ref{fig:ranking:layer:example}(b)) to compute the contrast. However, if every feature map is ranked (Fig. \ref{fig:ranking:layer:example}(c)), only one unique convolutional filter (Fig.~\ref{fig:ranking:layer:example}(d)) is needed  to compute the contrast. As a whole, a ranked feature map facilitates its \textit{statistical attributes} extraction. While the value of each feature, which is actually computed through classical convolutional or pooling operations, still reserves the \textit{structural attributes}.

We finally analyse the computational complexity of the ranking layer. The computational complexity of the forward propagation is $O(n\lg(n))$ serially and $O(\lg(n))$ parallelly, since it acctualy performs a sort operation. The computational complexity of the backward propagation is $O(n)$ serially and $O(1)$ parallelly, since it directly propagates the derivatives according to the correspondence $\mathcal{C}$.

\subsection{Learning Haze-relevant Features}
\label{sect:feature:learning}


Given the ranking layer and the CNN, three issues still need to be addressed to learn haze-relevant features, including: 1)~generating training data; 2) determining the structure of Ranking-CNN; 3) optimizing the parameters of Ranking-CNN.

Due to the lack of large-scale benchmarks, it is difficult to collect sufficient training data. Thus we address the first issue by generating massive synthesized hazy image patches for training the Ranking-CNN. As shown in Fig.~\ref{fig:Ranking-CNN:structure}, we first collect $400$ clear images from the Internet, including various types of scenes, such as mountain, forest, grass, city, building, street scene, etc. From these images, we randomly select $100,000$ clear image patches with the resolution $20\times 20$. Based on these patches, we follow the formation process of a hazy image in \eqref{eq:haze:formation} to generate massive hazy patches. Given a clear patch $\mathcal{B}$, we choose $10$ random transmission $t_\mathcal{B}$ between $(0,1]$ and assume that the transmission on each small image patch is constant. Thus the hazy patches can be synthesized via simulating the formation process of hazy images in \eqref{eq:haze:formation}. Since the main objective of Ranking-CNN is to learn haze-relevant features for transmission prediction, we use the same atmosphere light for all patches in the synthesization process ({\em i.e.}, $\left(1, 1, 1\right)^\text{T}$). Finally, we have $1,000,000$ synthesized hazy patches for learning haze-relevant features.

Before training the Ranking-CNN, we have to determine its structure. As shown in Fig. \ref{fig:Ranking-CNN:structure}, our Ranking-CNN has ten layers. The first layer is the input layer, which includes the RGB channels of a color image patch with resolution $20\times20$. The second layer is a convolutional layer, where the R, G, B maps are convolved with $5\times5$ convolutional kernels to generate $32$ feature maps with resolution $16\times16$. The third layer is a max pooling layer that sub-samples the input feature map over each $2\times2$ non-overlapping window. The fourth layer is the ranking layer, which operates separately on each input feature map. It sorts all the elements in an input feature map and outputs a ranked feature map with the same dimension. Note that the elements in the ranked feature map are in ascending order from left-top to right-bottom. The fifth layer is a convolutional layer, which includes 32 feature maps and the convolutional kernel size is $3\times 3$. The sixth layer is also a convolutional layer same with the fifth layer. The seventh layer is another max pooling layer which is the same as the third layer. After this layer, we finally obtain $32$ feature maps of size $2\times2$. The eighth layer, ninth layer (each with $64$ features) and the output layer (with $10$ output values) are all fully-connected layers. In our Ranking-CNN, we use rectified linear unit (ReLU) activation function \cite{Hinton:2010:ICML} for all convolutional layers and the first fully-connected layer. For each hazy patch $\mathcal{B}$,  the 10D output vector (denoted as $\textbf{Y}_\mathcal{B}$) are expected to approximate the label vector $\textbf{N}_\mathcal{B} = \left(n_\mathcal{B}^1, n_\mathcal{B}^2, \cdots, n_\mathcal{B}^{10}\right)^\text{T}$, where $n_\mathcal{B}^i\in\{0,1\}$ is a binary variable that can be calculated as
\begin{equation}
  n_\mathcal{B}^i = \left\{
    \begin{aligned}
      1,&~~\text{if}~t_\mathcal{B} \in \left(i/10 - 0.1, i/10\right] \\
      0,&~~\text{otherwise}
    \end{aligned}
  \right. .
\label{eq:n}
\end{equation}
In other words, we treat the Ranking-CNN as a multi-class classifier and try to optimize its parameters via maximizing the classification accuracy.

Intuitively, we can train an end-to-end network that predicts the transmission by replacing the output layer with a linear regression layer. However, since the output of the linear regression layer is only a variable varying between $(0,1]$, it is difficult to effectively train the deep network. To facilitate the training process, we adopt a two-stage training scheme. That is, we first convert the problem to a 10-category classification problem and train a Ranking-CNN model for classification. After that, the output layer is discarded and the output of the second fully-connected layer is used as features for training a random forest regressor to predict the transmission. In this manner, the training process of the Ranking-CNN is easier and the learned features, when they are combined with the random forest regressor, still have impressive performance in image dehazing.

To train the Ranking-CNN model, we minimize a soft-max loss function to optimize the parameters in the network. The loss function is defined as
\begin{equation}
\mathcal{L}\left(\textbf{N}_\mathcal{B}, \textbf{Y}_\mathcal{B}\right)
=-\log\left( \frac{e^{y_{\mathcal{B}}^j}}{\sum\limits_{i=1}^{10}e^{y_{\mathcal{B}}^i}} \right),
\label{eq:loss}
\end{equation}
where $y_\mathcal{B}^i$ is the $i$th element of $\textbf{Y}_\mathcal{B}$, and $j$ is an index that $n_\mathcal{B}^j = 1$.
To optimize the parameters in Ranking-CNN, we use the back-propagation algorithm with stochastic gradient descent solver \cite{LeCun:1998:IEEE}. We set the initial learning rate $r_{l_0}$ as $0.01$, the momentum as $0.9$, the mini-batch size as $64$. As shown in previous literatures \cite{Cai:2016:TIP, Ren:2016:ECCV}, it is helpful to decrease the learning rate along with the training process. Therefore, we update the learning rate as 
\begin{equation}
r_l=r_{l_0}\times\left(1+0.0001\times iter\right)^{-0.75},
\label{eq:rl}
\end{equation}
where $iter$ is the index of training iteration on each mini-batch. In the experiments, we perform $100$ epoches on the whole training data, and the 64D output of the second fully-connected layer are used for transmission prediction.

\subsection{Image Dehazing}
\label{sect:dehazing}
Based on the learned features, we further use the Random Forest \cite{Breiman:2001:ML} to learn a regression model between the transmission $t$ and the haze-relevant features. Our random forest model has $200$ trees and each tree random selects $1/3$ feature dimensions. For efficiency, we random select $1/100$ hazy image patches (\textit{i.e.}, $10,000$) to train the regression model. Note that we set the atmospheric light as a constant vector ({\em i.e.}, $\left(1,1,1\right)^\text{T}$) during training process. To relax this condition, we first apply white balance on the input image using our estimated atmospheric light $\textbf{A}$. In our approach, white balance is applied by dividing each channel $c$ of the input image by the corresponding channel of the estimated $\textbf{A}$ as 
\begin{equation}
  {\textbf{I}'}^{c}\left(x\right) = \frac{\textbf{I}^{c}\left(x\right)}{\textbf{A}^{c}\left(x\right)} = \frac{\textbf{J}^{c}\left(x\right)}{\textbf{A}^{c}\left(x\right)}t(x)+(1-t(x)).
\label{eq:white:balance}
\end{equation}
Thus $\textbf{I}'$ can be regarded to have atmospheric light of $(1,1,1)^\text{T}$ and the transmissions of $\textbf{I}'$ and $\textbf{I}$ are the same.

In the training data synthesizing process, we also assume that the transmission coefficients are locally consistent. However, we do not hold this assumption in the dehazing process. Therefore, we extract the haze-relevant features for every pixel in input image via selecting a $20\times 20$ patch centred at the pixel using the Ranking-CNN. With these features, the regression model is applied to estimate the transmission $t\left(x\right)$ for each pixel $x$. To avoid the artifacts near object edges, we further use guided filter \cite{He:2013:TPAMI} to smooth the initial estimated transmission for efficiency. Laplacian matting \cite{Levin:2008:TPAMI} also can be used instead to get more satisfactory results around edges. After obtaining the transmission $t(x)$ and atmosphere light $\textbf{A}(x)$ for each pixel $x$, we can dehaze the input image by applying the ideal dehazing process in \eqref{eq:dehaze}. Moreover, to avoid the strong fluctuation of recovered pixel when the transmission is very small, we set $t(x)=0.05$ if $t(x)<0.05$. Thus we can get the clear image as
\begin{equation}
  \textbf{J}\left(x\right) = \frac{\textbf{I}\left(x\right) - \textbf{A}\left(x\right)}{\textbf{max}\left(t\left(x\right), 0.05\right)} + \textbf{A}\left(x\right).
\label{eq:dehaze:AP}
\end{equation}

As the exposure is determined according to the hazy scene, the dehazed image usually tends to be underexposure, \textit{i.e.}, the luminance $\textbf{J}^l(x)$ of \textbf{J}(x) is usually much less than the luminance $\textbf{I}^l (x)$ of $\textbf{I}(x)$. Therefore, we adaptively increase the exposure as $\textbf{J}^*\left(x\right) = \lambda \textbf{J}\left(x\right)$, where $1 \leq \lambda \leq \frac{\sum_x \textbf{I}^l(x)}{\sum_x \textbf{J}^l(x)}$ is the exposure factor. As there are many regions which tend to be gray in the input hazy image, the dehazed image will be overexposure if $ \lambda = \frac{\sum_x \textbf{I}^l(x)}{\sum_x \textbf{J}^l(x)} $. As a compromise, the $\text{log}$ function is used in our method and
\begin{equation}
  \lambda = \text{log}\left(\frac{\sum_x \textbf{I}^l\left(x\right)}{\sum_x \textbf{J}^l\left(x\right)}\right) + 1.
\label{eq:dehaze:AP:R4}
\end{equation}
Then, the exposure can be increased and overexposure also can be avoided at the same time.

\section{Experiments}
\label{sect:experiments}

We first compare the dehazed results of our method and several previous methods on both synthetic and real benchmark images. Then we exploit the influence of the ranking layer and compare the features learned by our Ranking-CNN with previous haze-relevant features quantitatively.

\subsection{Comparisons with Previous Approaches}

\begin{table*}[t]
\caption{The $L_1$ errors on stereo dataset-syn. Left values indicate $L_1$ error in transmission. Right values indicate $L_1$ error in image.}
\label{table:l1:stereo1}
\center
\footnotesize
\tabcolsep 8pt 
\begin{tabular*}{156mm}{ c @{\hspace{3mm}} c @{\hspace{3mm}} c @{\hspace{3mm}} c @{\hspace{3mm}} c @{\hspace{3mm}} c @{\hspace{3mm}} c @{\hspace{3mm}} c }
\toprule
     & He {\em et al.} \cite{He:2009:CVPR} & Tang {\em et al.} \cite{Tang:2014:CVPR} & Zhu {\em et al.} \cite{Zhu:2015:TIP} & Berman {\em et al.} \cite{Berman:2016:CVPR} & Ren {\em et al.} \cite{Ren:2016:ECCV} & Cai {\em et al.} \cite{Cai:2016:TIP} & Ours \\
\midrule
  Aloe     &  0.100 / 0.191 & 0.060 / 0.087 & 0.175 / 0.141 & 0.060 / 0.086 & - / 0.195 & 0.089 / 0.096 & \textbf{0.051} / \textbf{0.070} \\ [0.5mm]
  Art      &  0.116 / 0.176 & 0.077 / 0.098 & 0.114 / 0.145 & 0.099 / 0.123 & - / 0.210 & 0.094 / 0.122 & \textbf{0.061} / \textbf{0.076} \\ [0.5mm]
  Barn     &  0.079 / 0.089 & 0.061 / 0.063 & 0.075 / 0.079 & 0.128 / \textbf{0.049} & - / 0.174 & 0.075 / 0.079 & \textbf{0.051} / 0.055 \\ [0.5mm]
  Bull     &  0.050 / 0.122 & 0.035 / 0.091 & 0.184 / 0.265 & 0.049 / 0.102 & - / 0.337 & 0.110 / 0.202 & \textbf{0.023} / \textbf{0.061} \\ [0.5mm]
  Cones    &  0.084 / 0.102 & 0.043 / 0.044 & 0.106 / 0.110 & 0.055 / 0.071 & - / 0.178 & 0.081 / 0.087 & \textbf{0.034} / \textbf{0.036} \\ [0.5mm]
  Dolls    &  0.061 / 0.110 & 0.038 / 0.069 & 0.152 / 0.201 & 0.067 / 0.095 & - / 0.272 & 0.076 / 0.132 & \textbf{0.032} / \textbf{0.060} \\ [0.5mm]
  Flower   &  0.059 / 0.105 & 0.046 / \textbf{0.066} & 0.146 / 0.172 & 0.066 / 0.145 & - / 0.239 & 0.098 / 0.135 & \textbf{0.045} / 0.068 \\ [0.5mm]
  Teddy    &  0.092 / 0.135 & 0.055 / \textbf{0.060} & 0.124 / 0.126 & 0.092 / 0.125 & - / 0.167 & 0.082 / 0.089 & \textbf{0.054} / 0.061 \\ [0.5mm]
  Tsukuba  &  0.068 / 0.093 & 0.077 / \textbf{0.123} & 0.173 / 0.253 & \textbf{0.060} / 0.113 & - / 0.329 & 0.117 / 0.182 &  0.077 / 0.125 \\ [0.5mm]
  Venus    &  0.042 / 0.074 & 0.046 / 0.103 & 0.159 / 0.239 & 0.051 / 0.163 & - / 0.310 & 0.114 / 0.196 & \textbf{0.035} / \textbf{0.079} \\ [0.5mm]
  \midrule
  Average  &  0.075 / 0.120 & 0.053 / 0.080 & 0.141 / 0.173 & 0.073 / 0.107 & - / 0.241 & 0.094 / 0.132 & \textbf{0.046} / \textbf{0.069} \\
\bottomrule
\end{tabular*}
\end{table*}

\begin{figure*}[!t]
  \centering
  \includegraphics[width = 180mm]{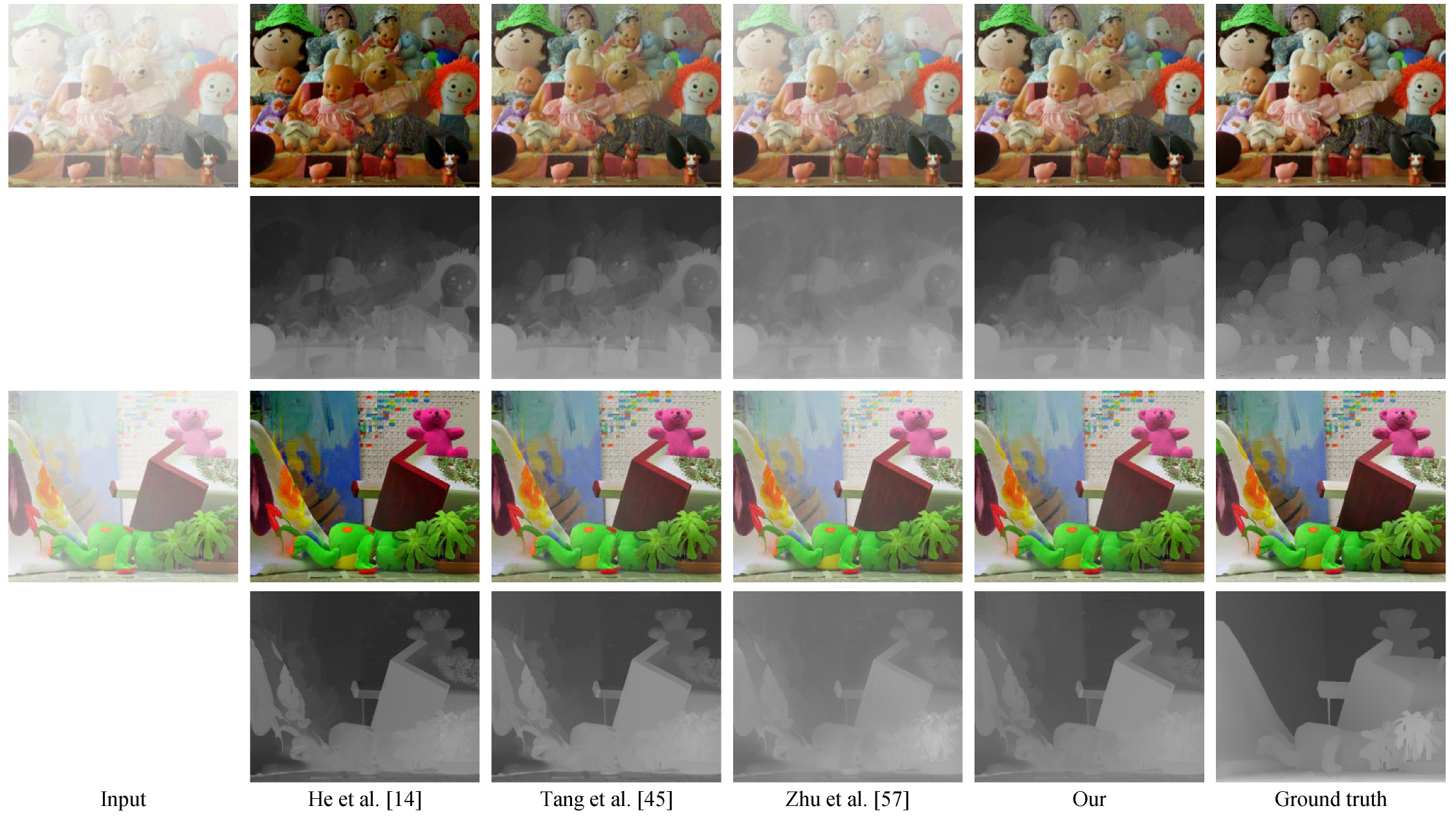}
\caption{Representative dehazed results on Dataset-Syn. We can see that, Zhu {\em et al.} \cite{Zhu:2015:TIP} usually under estimate the transmission, while \cite{He:2009:CVPR} and \cite{Tang:2014:CVPR} usually over estimate the haze, such as the light pink pig, the light brown heads, the red teddy and the gray areas.}
\label{fig:result:stereo1}
\end{figure*}

\begin{figure*}[!t]
  \centering
  \includegraphics[width = 174mm]{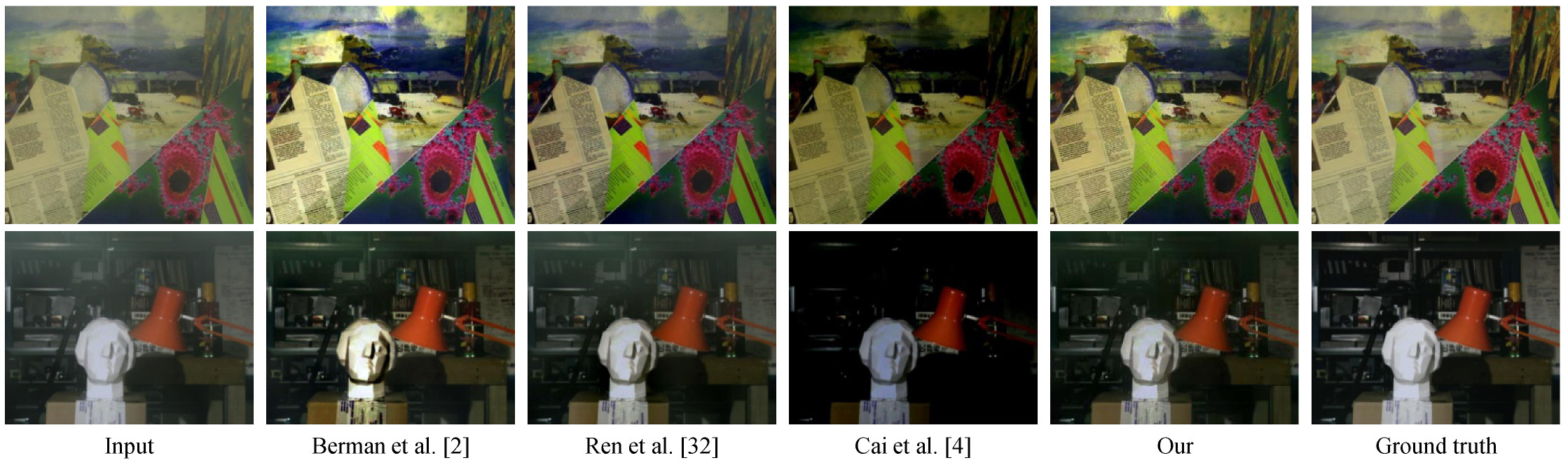}
\caption{Representative dehazed results on Dataset-Cap. We can see that our method can achieve satisfactory results on images with light haze. These results illustrate the robustness of our method.}
\label{fig:result:stereo2}
\end{figure*}

\begin{figure*}[!t]
  \centering
  \includegraphics[width = 168mm]{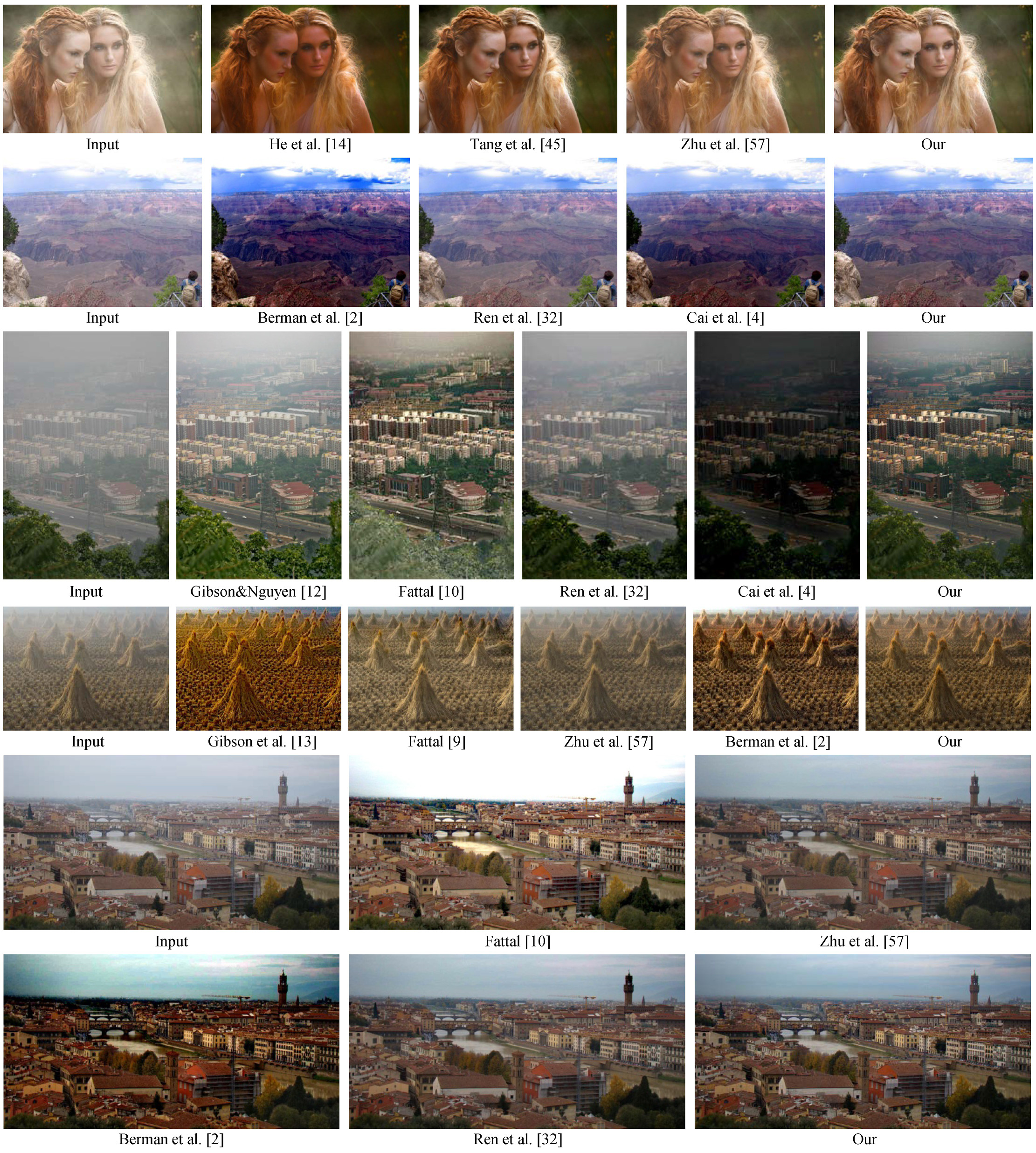}
\caption{Representative results obtained by our approach and previous methods. The results show that, our method can achieve visual better results on a lot of real benchmark images. Specially, our method suffers less over estimating problems or color shifts, such as the faces of the two actresses, and the green trees.}
\label{fig:result:visual}
\end{figure*}

The dehazed results and comparisons can be found in Fig.~\ref{fig:result:stereo1}, Fig.~\ref{fig:result:stereo2} and Fig.~\ref{fig:result:visual}, which are achieved respectively on synthetic hazy images with ground-truth clear images and transmissions, captured hazy images with known clear images, and real benchmark hazy images without ground-truth. The experimental results show that, our method can achieve better results compared with several previous methods both quantitatively and qualitatively. In our experiments, we implement our Ranking-CNN to learn haze-relevant features based on the open source deep learning framework Caffe \cite{Jia:2014:caffe}. We reimplement the methods of \cite{He:2009:CVPR} and \cite{Tang:2014:CVPR}, and directly use the published results or codes of other referenced methods, such as \cite{Fattal:2014:TOG, Zhu:2015:TIP, Berman:2016:CVPR, Ren:2016:ECCV, Cai:2016:TIP}. 

\begin{table*}[t]
\caption{The $L_1$ errors on Dataset-Cap. Values indicate $L_1$ error in image.}
\label{table:l1:stereo2}
\center
\footnotesize
\tabcolsep 8pt 
\begin{tabular*}{146mm}{ c @{\hspace{3mm}} c @{\hspace{3mm}} c @{\hspace{3mm}} c @{\hspace{3mm}} c @{\hspace{3mm}} c @{\hspace{3mm}} c @{\hspace{3mm}} c }
\toprule
     & He {\em et al.} \cite{He:2009:CVPR} & Tang {\em et al.} \cite{Tang:2014:CVPR} & Zhu {\em et al.} \cite{Zhu:2015:TIP} & Berman {\em et al.} \cite{Berman:2016:CVPR} & Ren {\em et al.} \cite{Ren:2016:ECCV} & Cai {\em et al.} \cite{Cai:2016:TIP} & Ours \\
\midrule
  Aloe     & 0.169 & 0.175 & \textbf{0.091} & 0.130 & 0.169 & 0.313 & 0.134 \\ [0.5mm]
  Art      & 0.136 & 0.087 & 0.073 & 0.090 & 0.079 & 0.231 & \textbf{0.064} \\ [0.5mm]
  Barn     & 0.054 & 0.046 & 0.070 & 0.087 & 0.061 & 0.117 & \textbf{0.041} \\ [0.5mm]
  Bull     & 0.064 & 0.075 & \textbf{0.046} & 0.065 & 0.087 & 0.206 & 0.053 \\ [0.5mm]
  Cones    & 0.093 & 0.064 & \textbf{0.053} & 0.057 & 0.104 & 0.213 & 0.057 \\ [0.5mm]
  Dolls    & 0.103 & 0.074 & 0.066 & 0.083 & 0.089 & 0.201 & \textbf{0.057} \\ [0.5mm]
  Flower   & 0.070 & 0.049 & 0.052 & 0.080 & 0.068 & 0.212 & \textbf{0.037} \\ [0.5mm]
  Teddy    & 0.126 & 0.108 & \textbf{0.067} & 0.141 & 0.129 & 0.197 & 0.089 \\ [0.5mm]
  Tsukuba  & 0.065 & 0.060 & 0.072 & 0.057 & 0.655 & 0.259 & \textbf{0.048} \\ [0.5mm]
  Venus    & \textbf{0.048} & 0.059 & 0.053 & 0.105 & 0.071 & 0.190 & 0.050 \\ [0.5mm]
  \midrule
  Average  & 0.093 & 0.080 & 0.064 & 0.089 & 0.092 & 0.214 & \textbf{0.063} \\
\bottomrule
\end{tabular*}
\end{table*}

In order to perform quantitative comparison, like some previous methods \cite{Tang:2014:CVPR, Ren:2016:ECCV, Cai:2016:TIP}, we synthesize ten hazy images based on stereo benchmark images published in \cite{Scharstein:2002:IJCV, Scharstein:2003:CVPR, Scharstein:2007:CVPR}, which is denoted as Dataset-Syn. Each image from this dataset has two types of ground-truth, including a haze-free image and a ground-truth transmission map. To be fair, we follow the experiments set-up in \cite{Tang:2014:CVPR} and set the transmission $t(x)=0.8\times d(x)$ for each pixel $x$, where $d(x)$ is the disparity. Table~\ref{table:l1:stereo1} shows the $L_1$ error comparisons in transmission and image of our method and \cite{He:2009:CVPR, Tang:2014:CVPR, Zhu:2015:TIP, Berman:2016:CVPR, Ren:2016:ECCV, Cai:2016:TIP}. The $L_1$ error in transmission is calculated between the estimated and ground-truth transmission maps, and the $L_1$ error in image is calculated between the dehazed and haze-free images. Overall, as can be seen, our method achieves the best results and has over $10\%$ lower average $L_1$ error in estimated transmission and dehazed image compared with these methods. There are two dehazed results illustrated in Fig.~\ref{fig:result:stereo1}, we can see that \cite{He:2009:CVPR} usually over estimate the haze, such as the light pink pig, the light brown heads, the red teddy and the gray areas. \cite{Tang:2014:CVPR} also suffers this problem as the multi-scale dark channel features are the most important features in their method. On the contrary, our method suffers less over estimated problems.

Though the Dataset-Syn are synthesized following the abstractly formulation of hazy images \eqref{eq:haze:formation}, however, the physical process may not follow it precisely. To this end, inspired by the construction of image matting benchmark \cite{Rhemann:2008:CVPR, Rhemann:2009:CVPR}, we design a process to directly capture a hazy image as well as its corresponding clear version. We first use a Lenovo 22'' monitor to display each clear image in Dataset-Syn, and capture it by a Cannon 650D DSLR camera. After that, an ultrasonic humidifier is used to fill vapour between the monitor and the camera. Then the camera captures the hazy image with all the other settings and parameters unchanged. Except the vapour, the environment settings and camera parameters are unchanged, therefore the captured clear image can be regarded as the ground-truth of its corresponding captured hazy image. This captured dataset is denoted as Dataset-Cap. Table~\ref{table:l1:stereo2} shows the $L_1$ error comparisons in image of our method and \cite{He:2009:CVPR, Tang:2014:CVPR, Zhu:2015:TIP, Berman:2016:CVPR, Ren:2016:ECCV, Cai:2016:TIP}. Our method still achieves the best results. There are two dehazed results illustrated in Fig.~\ref{fig:result:stereo2}. As Fig.~\ref{fig:result:stereo1} and Fig.~\ref{fig:result:stereo2} show, the hazy of Dataset-Syn is dense and it of Dataset-Cap is light. Our method can achieve the best performance on both these two datasets, which means that our method performs robust under different hazy density than other methods.

Finally, we conduct a subjective test to visually compare the results of our method, \cite{He:2009:CVPR}, \cite{Tang:2014:CVPR}, \cite{Fattal:2014:TOG}, \cite{Berman:2016:CVPR}, \cite{Cai:2016:TIP} and \cite{Ren:2016:ECCV} on $69$ benchmark images. For a fair comparison, we use the results which are published by \cite{He:2009:CVPR}, \cite{Tang:2014:CVPR}, \cite{Fattal:2014:TOG}, and generate the results using the codes which are published by \cite{Berman:2016:CVPR}, \cite{Cai:2016:TIP}, \cite{Ren:2016:ECCV}. Since each paper only publishes results on a portion of the $69$ images, we obtain $1012$ pairs of dehazed results in total. Fifteen subjects are invited to perform this experiment. All these subjects have normal or corrected normal visual acuity and normal color vision. The results are shown on a normal $22''$ display with $1680 \times 1050$ resolution. The display is placed in a room with fluorescent lamps. On each of such pair-wise comparisons, four images are shown in a $2 \times 2$ grid. The top-left is the input hazy image. The top-right is the ground truth hazy free image (if it exists, otherwise the input hazy image). The bottom-left and bottom-right are the dehazed results from two methods, each result is random shown in left or right. Each image is shown with no more than $640 \times 480$ resolution, which also can be shown with its original resolution in a new window by click. Each subject is requested to observe each comparison and determine which dehazed result is better. Averagely, each subject takes about $75$ minutes to perform the test. Note that the methods that are being compared are blind to the subjects. Among all these $1012 \times 16 = 16192$ pair-wise comparisons, our method achieves the first place and outperforms the other methods for $3221$ times, while \cite{He:2009:CVPR} takes the second place ($2649$ times). These results, together with the objective performance, indicate that our method performs the best in both objective and subjective experiments compared with the several referenced methods.

We also show some dehazed results on real world images in Fig.~\ref{fig:result:visual}. It shows that, our method can achieve visual better results on a lot of real benchmark images. Specially, our method suffers less over estimating problems and color shifts, such as the faces of the actresses and the green trees.

\subsection{Performance Analysis}

Beyond the performance comparisons, in this section we conduct a number of small experiments to validate the performance of our approach from multiple perspectives. For quantitatively evaluation, we further generate $400,000$ hazy patches with synthetic transmission as validation set.

\begin{figure}[!t]
  \centering
  \includegraphics[width = 84mm]{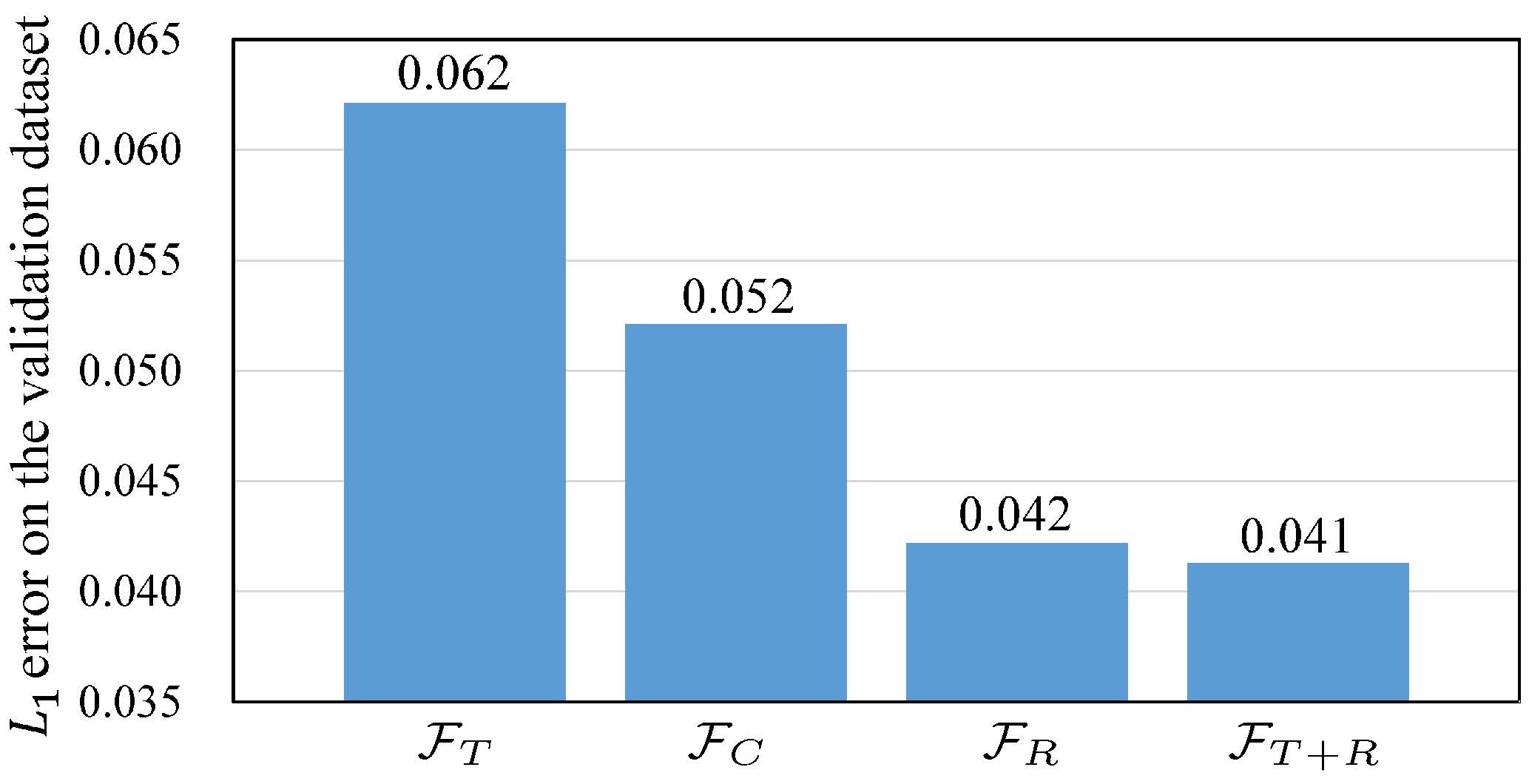}
\caption{The $L_1$ errors on validation set when different features are used for dehazing. $\mathcal{F}_{T}$: features used in \cite{Tang:2014:CVPR}; $\mathcal{F}_{C}$: features learned by the classical CNN; $\mathcal{F}_{R}$: features learned by the Ranking-CNN; $\mathcal{F}_{T + R}$: the combination of $\mathcal{F}_{T}$ and $\mathcal{F}_{R}$.}
\label{fig:mse:patches}
\end{figure}

\textbf{Features comparison}. In the first experiment, we compare the performance of various types of features, including the 64D features learned by the Ranking-CNN (denoted as $\mathcal{F}_R$), the 64D features learned by the classical CNN (removing the ranking layer in the Ranking-CNN, denoted as $\mathcal{F}_C$), the 325D features designed by \cite{Tang:2014:CVPR} (denoted as $\mathcal{F}_T$) and the combination of $\mathcal{F}_T$ and $\mathcal{F}_R$ (denoted as $\mathcal{F}_{T+R}$). Note that the 325D features of $\mathcal{F}_T$ consist of multi-scale dark channel priors and local max contrasts, hue disparity and multi-scale local max saturation. For efficiency issue, we random select $1/100$ training patches (\textit{i.e.}, $10,000$) from that are used by the Ranking-CNN to train the random forest regression model. Figure~\ref{fig:mse:patches} shows the $L_1$ error in transmission on validation set using different combinations of the features, \textit{i.e.} $\mathcal{F}_{T}$, $\mathcal{F}_{C}$, $\mathcal{F}_{R}$ and $\mathcal{F}_{T+R}$. We can see that the features from the Ranking-CNN outperform those from \cite{Tang:2014:CVPR} by $32\%$ in terms of $L_1$ error. Moreover, our Ranking-CNN features achieve about $20\%$ better compared with the classical CNN features. If we combine our Ranking-CNN features with the features used by \cite{Tang:2014:CVPR}, the $L_1$ error is only decreased slightly ($0.001$), which means that our Ranking-CNN features not only capture most information in the previous hand-crafted features, but also learn more information from the massive data automatically. This experiment also shows that both structural features ({\em e.g.}, CNN features) and statistical features ({\em e.g.}, features used in \cite{Tang:2014:CVPR}) are useful for transmission estimation.

\begin{figure}[!t]
  \centering
  \includegraphics[width = 84mm]{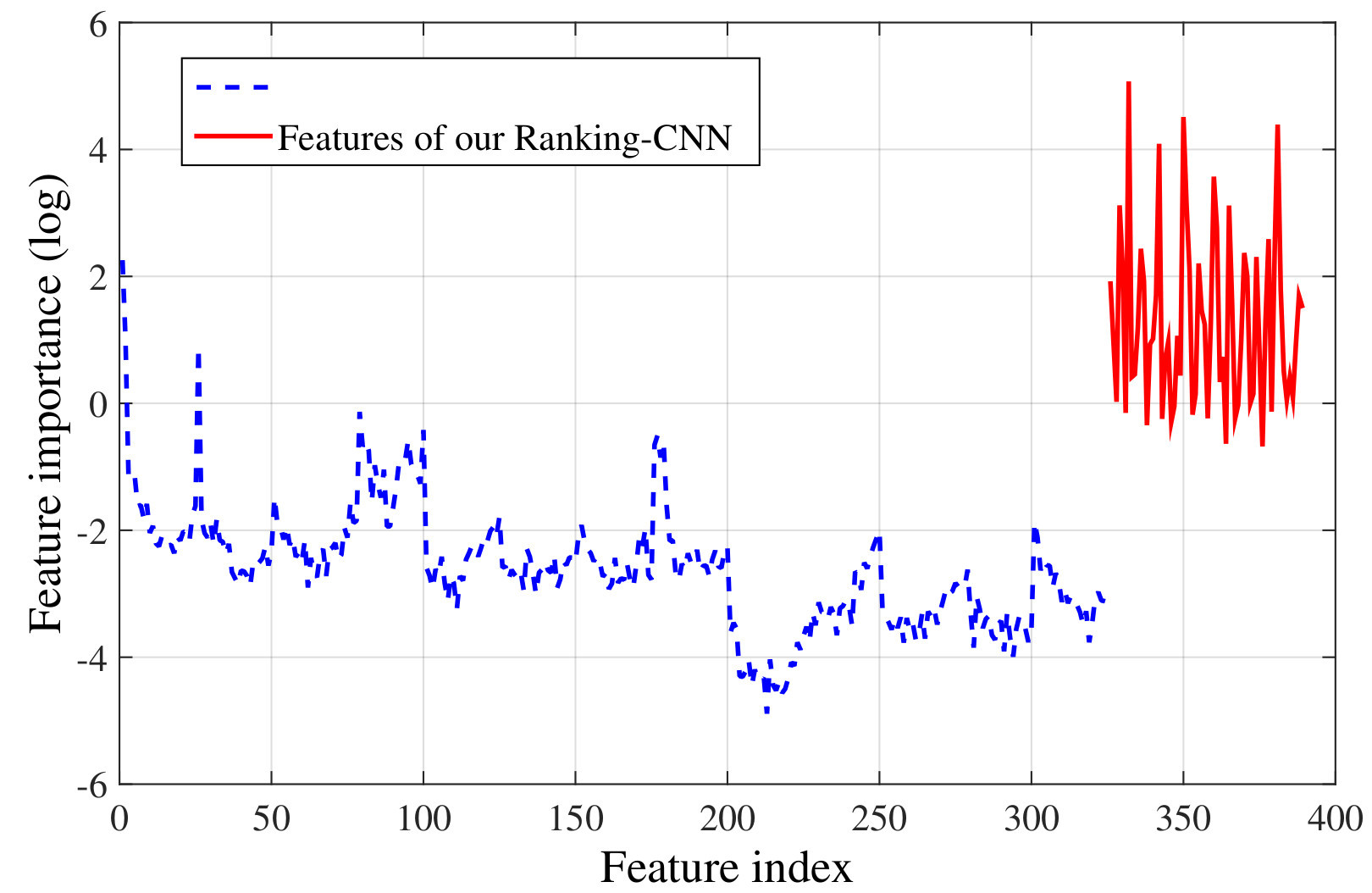}
  \put(-191,138){\scriptsize Features of Tang \textit{et al.} \cite{Tang:2014:CVPR}}
\caption{The importance of features learned from Ranking-CNN and features heuristically designed in \cite{Tang:2014:CVPR}.}
\label{fig:features:importance}
\end{figure}

We further explore the importance of each dimension in features $\mathcal{F}_{T+R}$, which consists of the 64D features from the Ranking-CNN and 325D features used in \cite{Tang:2014:CVPR}. All these features are incorporated to train a random forest regressor, and the importance of each feature dimension can be obtained. As illustrated in Fig.~\ref{fig:features:importance}, we plot the importance of each feature dimension which can be obtained from the trained random forest regressor. It is obviously that our Ranking-CNN features are more important than the previous features used in \cite{Tang:2014:CVPR}. Moreover, the sum importance of the Ranking-CNN features is $708.48$, while the sum importance of the previous features $\mathcal{F}_{T}$ is only $47.41$, which shows that the Ranking-CNN features are powerful and remarkably outperform the previous heuristic designed features used in \cite{Tang:2014:CVPR}.

We also compare the features generated from different layers by training the regression model. The $L_1$ errors on the same validation dataset are $0.050$, $0.043$ and $0.042$ by using the 128D features generated from the second pooling layer, the 64D features generated from the first fully-connected layer and the 64D features generated from the second fully-connected layer, respectively. This may imply that the features from deeper layers are more powerful. Thus we adopt the 64D features generated from the second fully-connected layer for transmission estimation.

\begin{figure}[t]
  \centering
  \includegraphics[width = 84 mm]{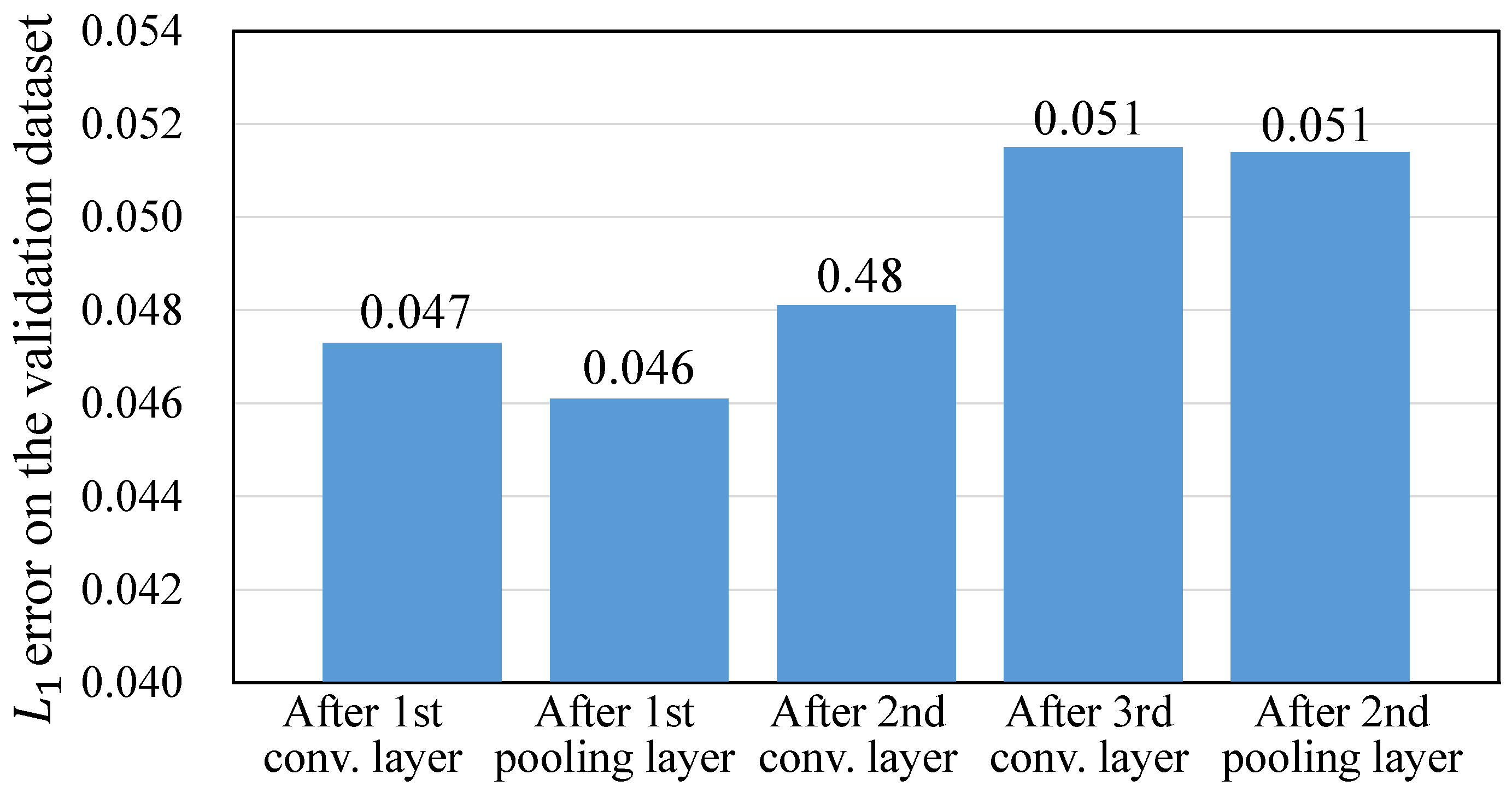}
  \caption{The $L_1$ error in transmission on the validation set using our Ranking-CNN when the ranking layer is placed at different locations.}
\label{fig:ranking:place}
\end{figure}

\begin{figure}[t]
  \centering
  \includegraphics[width = 84 mm]{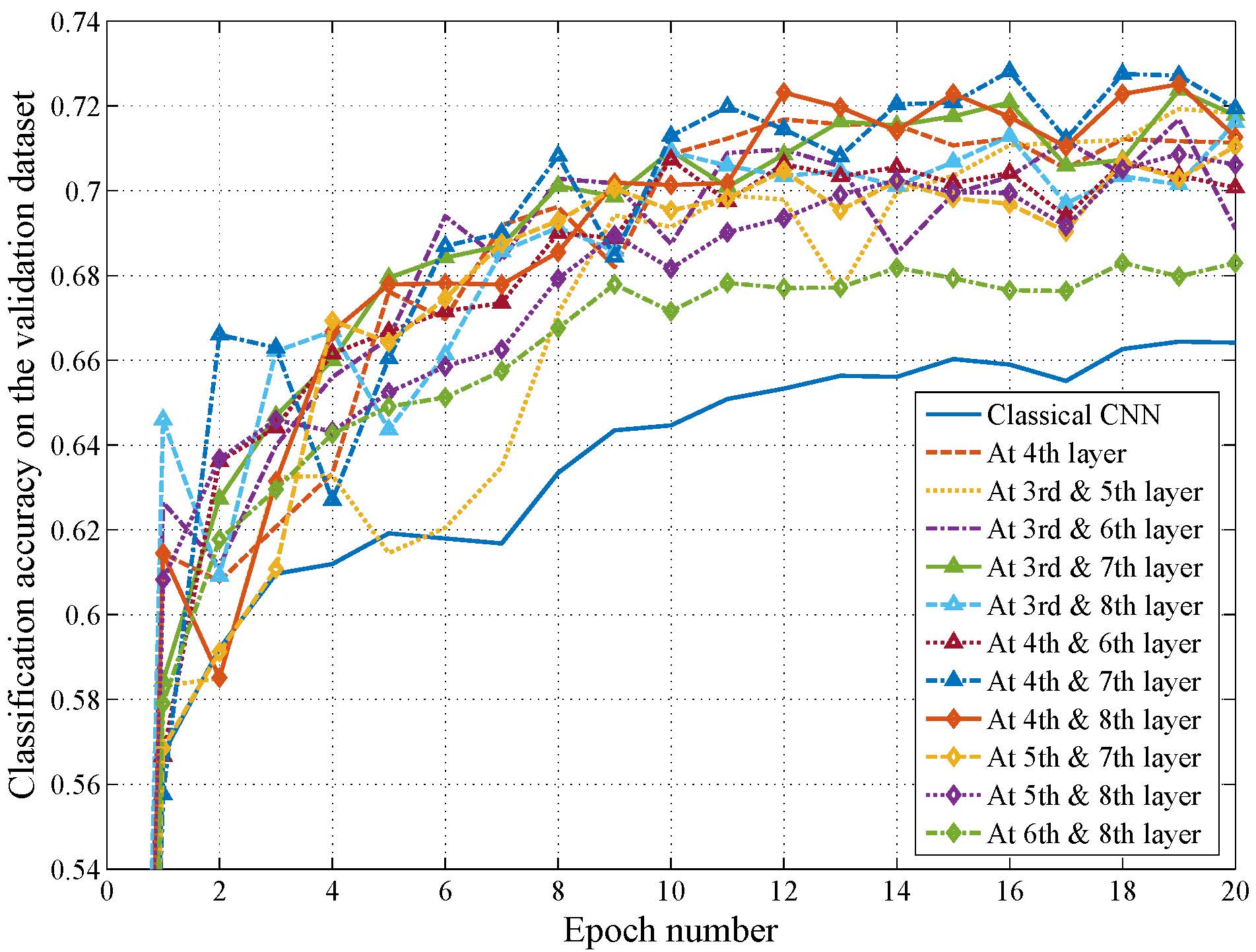}
  \caption{The classification accuracy on the validation dataset when two ranking layers are placed at all possible locations in CNN.}
\label{fig:2rankinglayers}
\end{figure}

\textbf{The location of the ranking layer}. In the third experiment, we show the performance of the Ranking-CNN when the ranking layer is placed at different locations. In the experiment, the ranking layer is placed after the first convolutional layer, the first pooling layer (as in Fig. \ref{fig:Ranking-CNN:structure}), the second and third convolutional layer, and the second pooling layer respectively. Figure~\ref{fig:ranking:place} shows the $L_1$ error in transmission on the validation set respectively after $20$ training epoches. We can see that when the ranking layer is placed at the fourth layer (after 1st pooling layer), our dehazing model achieves the minimal $L_1$ error. To explain this phenomena, we rethink the problem from another perspective: what features will be extracted without the ranking layer? As state in \cite{Zeiler:2014:ECCV}, the shallow layers of the classical CNN extract low-level features like boundaries and contrasts, while the deep layers extract high-level features like patterns and objects. In existing studies haze has been proved to be tightly correlated with low-level structural features like boundaries \cite{Meng:2013:ICCV} as well as the statistical information like dark channel prior \cite{He:2009:CVPR} and color-lines prior \cite{Fattal:2014:TOG}. By inserting the ranking layer after the first pooling layer, the proposed deep network can make full use of the low-level structural features like boundaries, while additional statistical information are also incorporated by using the ranking operations. By fusing both the low-level structural information and statistical information, the proposed method can achieve the best performance by placing the ranking layer at the fourth layer.

In the fourth experiment, we exploit the performance of the Ranking-CNN that uses two ranking layers. The two ranking layers are placed at all possible locations in the CNN. As shown in Fig.~\ref{fig:2rankinglayers}, the best performance is achieved by placing one ranking layer at the fourth layer ({\em i.e.}, after the first pooling layer) and the other one at the seventh layer ({\em i.e.}, after the second convolutional layer). However, the performance improvement, compared with the Ranking-CNN with only one ranking layer, is marginal (about $0.8\%$ on the validation dataset, after $20$ epoches). Considering the additional computational cost in the ranking layer, we still adopt one ranking layer for image dehazing.

\begin{figure}[t]
  \centering
  \includegraphics[width = 50 mm]{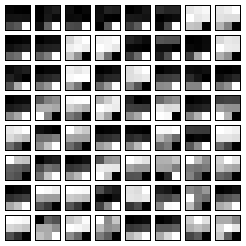}
\caption{The visualization of 64 filters randomly sampled from the second convolutional layer of the Ranking-CNN.}
\label{fig:filters}
\end{figure}

\textbf{Visualize the network}. In the fifth experiment, we try to explain why the features learned by the Ranking-CNN are useful for image dehazing. We randomly select and visualize $64$ filters from the second convolutional layer in Fig.~\ref{fig:filters}. We can see that these filters actually provide cues on which elements in a local patch of a feature map should be referred to in extracting haze-relevant features. For instance, the filter at the left-top corner may imply that the largest value in a local patch should be considered for extracting haze-relevant features. It is somehow similar to the mechanism of dark channel prior, while the main difference is that various types of haze-relevant features are extracted by referring to different combinations of elements in a local patch. In this manner, the Ranking-CNN extracts an over-complete set of haze-relevant features, which are then weighted and selected in the random forest regressor. In this manner, Ranking-CNN demonstrates impressive performance in dehazing images.

\begin{figure}[t]
  \centering
  \includegraphics[width = 88 mm]{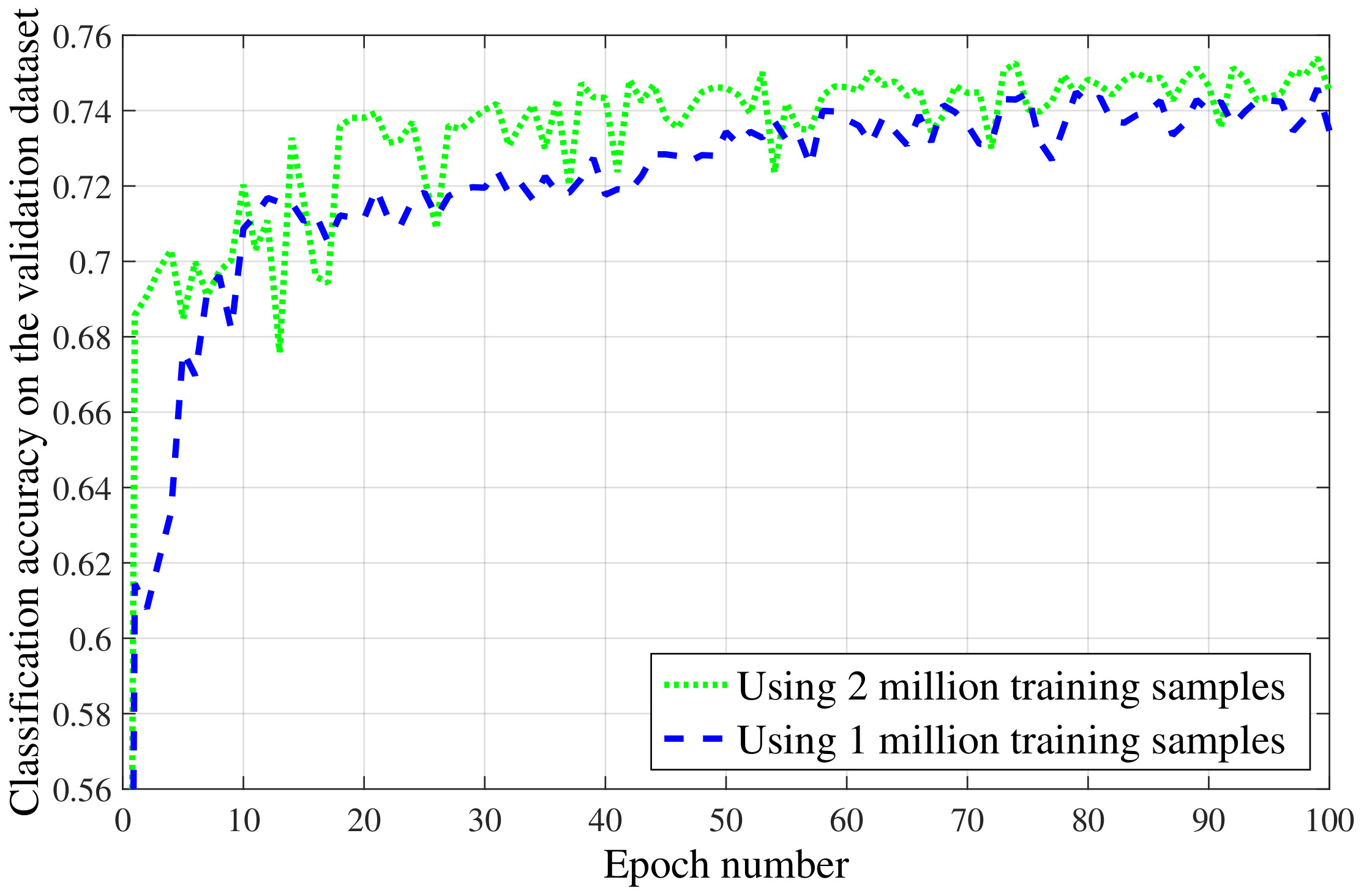}
  \caption{The classification accuracy of two Ranking-CNN models on the same validation set, which are trained on one million or two million synthetic training samples, respectively. We can see that more training data generally bring better performance.}
\label{fig:more:data}
\end{figure}

\textbf{The size of training data}. In the sixth experiment, we explore the influence when different numbers of synthetic training data are used in training the Ranking-CNN model. As shown in Fig.~\ref{fig:more:data}, the accuracy of the Ranking-CNN on the same validation set increases about $1\%$ after $100$ epoches when $2$ million training samples are used, while the training time is doubled as well. This result implies that our proposed method still has potential to be further improved by simply generating more synthetic training data. Considering the efficiency in the training stage, we use 1 million training samples in all the other experiments.

\begin{figure}[t]
  \centering
  \includegraphics[width = 88 mm]{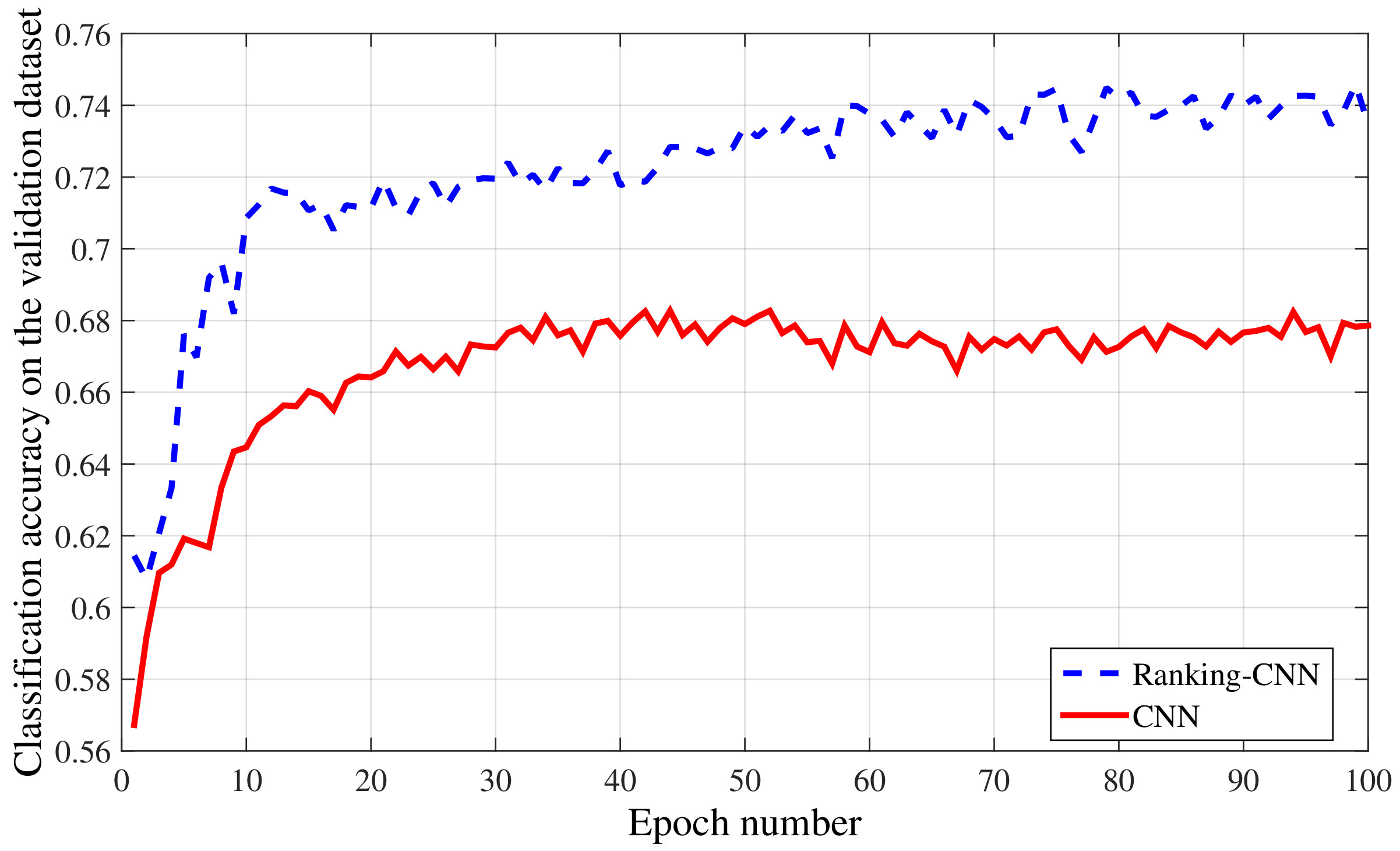}
  \caption{The classification accuracy of our Ranking-CNN when different number of epoches are performed in the training process. Our Ranking-CNN can converge as quickly as classical CNN.}
\label{fig:testing:error}
\end{figure}

\textbf{The convergence speed}. In the seventh experiment, we compare the convergence speeds between the Ranking-CNN and the classical CNN. As shown in Fig.~\ref{fig:testing:error}, the convergence speed of the Ranking-CNN is comparable to the classical CNN. This may be caused by the fact that in the ranking layer the partial derivatives of the loss function with respect to each output feature can be directly passed to its corresponding input feature. Since the ranking layer is parameter-free, adding a ranking layer will not dramatically increase the difficulties in training the network.

\textbf{Different regressors}. In the second experiment, we test the performance of different types of regressors. Besides the random forest, we select three other regressors, including linear regressor, logistic regressor and SVM regressor (with radial basis function kernel). The $L_1$ errors of these regressors on the same validation dataset are $0.042$ (random forest), $0.057$ (linear), $0.054$ (logistic) and $0.060$ (SVM). As random forest regressor achieves the best performance, we employ it in transmission estimation.

\textbf{The end-to-end method}. To explore the performance of the end-to-end method, we replace the output layer of our Ranking-CNN by a linear regression layer. Then the modified Ranking-CNN can directly predict the transmission. However, though it is actually more efficient, the performance is unsatisfactory. In fact, its mean $L_1$ error on test data is $0.073$, while our method achieves $0.042$. The reason may be that, when we take the network as a regressor and train it to predict the transmission, the $L_2$ loss function is a common choice, which is also used in our experiment. When we train a classification network, we use the soft-max loss function. We can see that, the soft-max loss function is steeper than the $L_2$ loss function, which means that the classification network can be updated more effectively. Moreover, since the classification network outputs a number of probabilities about each label other than a real value, the learned features tend to be more various.

\textbf{Running time}. The last experiment is about the Running time. Our experiments are performed on a 3.1GHz PC with a NVIDIA Gerforce GTX980 GPU. Our feature learning and extracting algorithm is implemented based on Caffe. It takes about $400$ seconds to perform one training epoch on all $1$ million training samples using GPU. In feature extracting process, it takes about $283$ seconds to extract features for $1$ million patches, while classical CNN takes about $247$ seconds. We use a C implementation of random forest and it takes about two minutes to train the regression model on our $10,000$ training samples using CPU, and takes about $0.25$ seconds to predict initial transmission of $10,000$ patches. The other parts of our method are implemented using matlab, which takes several seconds for a typically $640 \times 480$ image. Our method achieve satisfactory performance quantitatively and qualitatively, the weakness is its efficiency. Compared with several previous methods, our method takes more time. The main reason is that we extract features and estimate the transmission for every pixel according its local patch. However, as the transmissions are correlated in a local patch, we can simultaneously estimate the transmissions of more pixels in the future work. Then, the running time can be decreased more than one order of magnitude.

\section{Conclusion}
\label{sect:conclusion}
This paper presents a method to dehaze an image based on the features which are automatically learned from massive hazy images. To this end, a novel ranking layer is proposed to form the Ranking-CNN, that can learn haze-relevant features more effectively compared with the classical CNN. Equipped with the novel ranking layer, our Ranking-CNN can capture the structural and statistical features simultaneously. Based on the learned features, a regression model is further trained to predict haze density for effective haze removal. Experimental results show that our Ranking-CNN features are effective. The proposed image dehazing method, which is based on the features, also achieves satisfactory results on synthetic and real world data. At the same time, as we extract features for every pixel, the weakness of our method is its efficiency, which should be further improved in the future work, \textit{i.e.}, via adopting FCN framework \cite{Long_2015_CVPR} to reduce redundant computations. 



\bibliographystyle{abbrv}
\bibliography{dehaze_ddnn_bib}

\balance
\end{document}